\documentclass[lettersize,journal]{IEEEtran}
\usepackage{amsmath,amsfonts}
\usepackage{algorithmic}
\usepackage{algorithm}
\usepackage{array}
\usepackage[caption=false,font=footnotesize,labelfont=rm,textfont=rm]{subfig}
\usepackage{textcomp}
\usepackage{stfloats}
\usepackage{url}
\usepackage{verbatim}
\usepackage{graphicx}
\usepackage{cite}
\usepackage{bm}
\usepackage{bbm}
\usepackage{dsfont}
\usepackage{multirow}
\usepackage{microtype}
\usepackage{booktabs}
\usepackage{array}
\usepackage{makecell}
\usepackage{microtype}
\usepackage{booktabs}
\usepackage{array}
\usepackage{makecell}
\usepackage{xcolor}
\usepackage{wrapfig}
\usepackage{soul}
\usepackage{thmtools}
\usepackage{thm-restate}
\usepackage{flexisym}
\usepackage{tikz,xcolor}
\usepackage[implicit=false]{hyperref}

\hypersetup{hidelinks,
	colorlinks=true,
	allcolors=black,
	pdfstartview=Fit,
	breaklinks=true}

\definecolor{lime}{HTML}{A6CE39}
\DeclareRobustCommand{\orcidicon}{
\begin{tikzpicture}
\draw[lime, fill=lime] (0,0)
circle[radius=0.16]
node[white]{{\fontfamily{qag}\selectfont \tiny \.{I}D}};
\end{tikzpicture}
\hspace{-2mm}
}
\foreach \x in {A, ..., Z}{%
\expandafter\xdef\csname orcid\x\endcsname{\noexpand\href{https://orcid.org/\csname orcidauthor\x\endcsname}{\noexpand\orcidicon}}
}

\begin{document}

\title{Consistent Assistant Domains Transformer for Source-free Domain Adaptation}

\author{Renrong Shao\orcidA{},
        Wei Zhang\orcidB{},
        Kangyang Luo\orcidC{},
        Qin Li\orcidD{},
        and Jun Wang\orcidE{} 
\thanks{
Manuscript received 16 December 2024; revised 2 June and 14 August 2025; accepted 10 September 2025. date of current version 14 September 2025.
This work was supported in part by Key Laboratory of Advanced Theory and Application in Statistics and Data Science, Ministry of Education. 
The associate editor coordinating the review of this manuscript and approving it for publication was Dr. Jochen Lang. (Corresponding author: Wei Zhang, Qin Li, Jun Wang)

Renrong Shao is with the Faculty of Military Health Services, Naval Medical University (Second Military Medical University), Shanghai 200433, China (e-mail: roryshaw6613@smmu.edu.cn)

Kangyang Luo is with the Department of Computer Science and Technology, Tsinghua University, Beijing 100084, China. (e-mail: luoky@mail.tsinghua.edu.cn)

Qin Li is with the Software Engineering Institute, East China Normal University, Shanghai 200262, China (e-mail: qli@sei.ecnu.edu.cn). 

Wei Zhang and Jun Wang are with the School of Computer Science and Technology, East China Normal University, Shanghai 200262, China (e-mail: zhangwei.ltt@gmail.com, wongjun@gmail.com).
}

}

\markboth{IEEE TRANSACTIONS ON IMAGE PROCESSING, ~Vol.~, ~2025, DOI 10.1109/TIP.2025.3611799}%
{Shell \MakeLowercase{\textit{et al.}}: A Sample Article Using IEEEtran.cls for IEEE Journals}




\maketitle

\begin{abstract}
Source-free domain adaptation (SFDA) aims to address the challenge of adapting to a target domain without accessing the source domain directly. However, due to the inaccessibility of source domain data, deterministic invariable features cannot be obtained. Current mainstream methods primarily focus on evaluating invariant features in the target domain that closely resemble those in the source domain, subsequently aligning the target domain with the source domain. However, these methods are susceptible to hard samples and influenced by domain bias. In this paper, we propose a {\textbf{C}onsistent \textbf{A}ssistant \textbf{D}omains \textbf{Trans}former for SFDA, abbreviated as \textbf{CADTrans}}, which solves the issue by constructing invariable feature representations of domain consistency. Concretely, we develop an assistant domain module for CADTrans to obtain diversified representations from the intermediate aggregated global attentions, which addresses the limitation of existing methods in adequately representing diversity. Based on assistant and target domains, invariable feature representations are obtained by multiple consistent strategies, which can be used to distinguish easy and hard samples. Finally, to align the hard samples to the corresponding easy samples, we construct a conditional multi-kernel max mean discrepancy (CMK-MMD) strategy to distinguish between samples of the same category and those of different categories. Extensive experiments are conducted on various benchmarks such as Office-31, Office-Home, VISDA-C, and {DomainNet-126}, proving the significant performance improvements achieved by our proposed approaches. 
{~\textit{Code is available at https://github.com/RoryShao/CADTrans.git.}} 
\end{abstract}


\section{Introduction}
\IEEEPARstart{D}{eep} learning has achieved remarkable success in various applications of computer vision but also suffers from some drawbacks~\cite{russakovsky2015imagenet,yang2022image}.
Firstly, the success of deep learning relies on huge manual annotation, which is extremely expensive in real-world applications.
Besides, owing to insufficient generalization, applying existing models to other relevant scenarios usually encounters the domain shift.
To address the said issue, unsupervised domain adaptation~(UDA) has been introduced, which aims to transfer knowledge from the source domain to the target domain.  The key idea of UDA is to exploit the network to extract the labeled source domain and unlabeled target domain feature, and then project the feature in common feature space to learn domain invariant~\cite{pan2009survey,ganin2015unsupervised,long2015learning,su2024m} and common semantic information~\cite{goodfellow2014generative,long2018conditional,su2021few}.
Nevertheless, since security privacy protection and data transmission limitations, the model cannot access the source data directly in some scenarios.
\begin{figure}[t!]
\centering 
\includegraphics[width=0.9\linewidth]{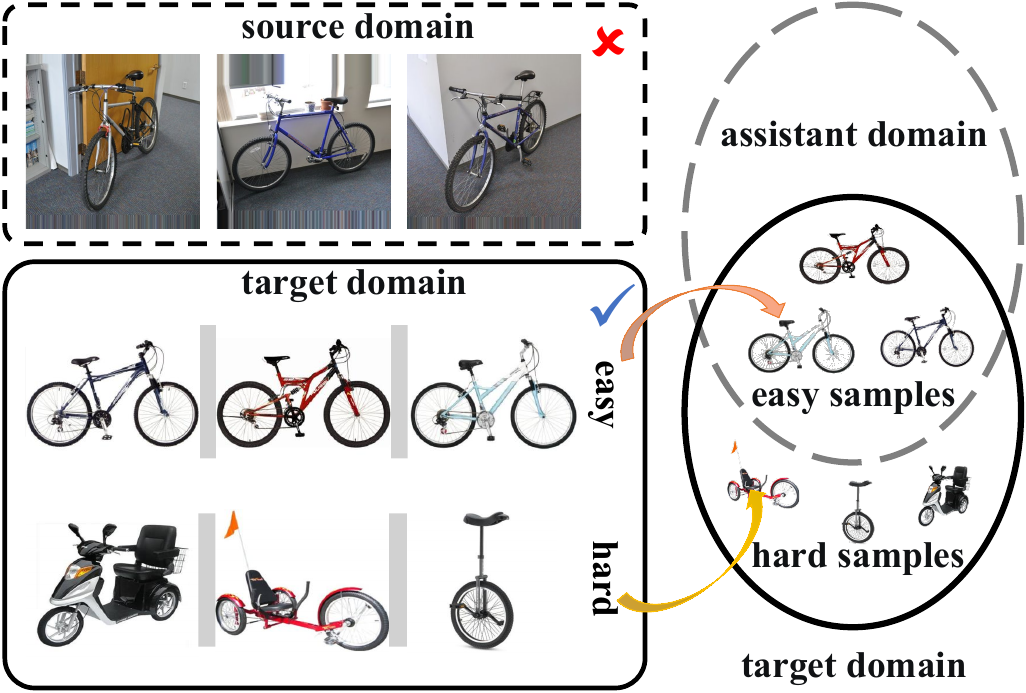}
\vspace{-2mm}
\caption{Principle of our proposed methodology. The \textbf{Left}: Easy samples of target domain are more similar to the source data, while hard samples have great discrepancies. The \textbf{Right}: 
Construct an assistant domain to obtain invariant features by domain consistency strategies to evaluate easy samples and align hard samples.}
\vspace{-4mm}
\label{fig:overview}
\end{figure}

To overcome the data-absent problem, some prospective efforts~\cite{saito2018maximum,li2020model,xia2021adaptive,liang2020we,yang2021generalized,hou2016unsupervised,ren2022multi} work on source-free domain adaptation~(SFDA). In general, these methods roughly fall into two branches: adversarial-based approaches and pseudo-label evaluation. Adversarial-based approaches mainly utilize the generator to synthesize the target feature distribution and aim to improve the generalization performance of the target domain~\cite{saito2018maximum,li2020model,xia2021adaptive}. However, these approaches incur extensive resources and time to optimize the generator and synthetic images, and the generated samples cannot fully reflect the distribution of the real data when encountering complex data scenarios, thereby limiting the generalization performance. While pseudo-label evaluation approaches commonly leverage the pre-trained model of the source domain to evaluate the centroid by clustering the nearest neighborhood feature, and mitigate variation between cross-domains by certain constraints~\cite{hou2016unsupervised,liang2020we,yang2021generalized,ren2022multi}. Although this method has achieved promising results, it is still vulnerable to erroneous evaluations caused by hard samples.  For example, in the Office31 dataset as the \textbf{Left} of Fig.~\ref{fig:overview}, easy samples are closer to the source domain while hard samples have great discrepancies.
These hard samples are difficult to align and are more prone to produce incorrect evaluations, which further exacerbates domain shift~\cite{liang2020we, zhang2022divide}.

Moreover, due to the inability to access source domain data, the primary challenge in SFDA lies in the difficulty of obtaining feature representations from the source domain directly. 
This limitation hampers our ability to align the source and target domains through invariant feature representations, i.e., intermediate features whose distribution is the same in source and target domains, while at the same time achieving small error on the source domain~\cite{zhao2019learning}.
This situation leads us to consider the following question: \textit{Can we distinguish invariant feature representations by employing specific strategies if we derive diverse representations from the reconstructed source domain?} Our analysis of the datasets revealed that the target domain, given its complexity, can be categorized into easy samples that resemble those of the source domain and more challenging hard samples that are unique to the target domain. Consequently, this insight motivates us to address this gap by constructing a source-like assistant domain, as illustrated in the \textbf{Right} of Fig.~\ref{fig:overview}. This approach aims to secure invariant feature representations through consistent strategies while effectively evaluating easy samples.

\begin{figure*}[!ht]
  \centering
   \includegraphics[width=\linewidth]{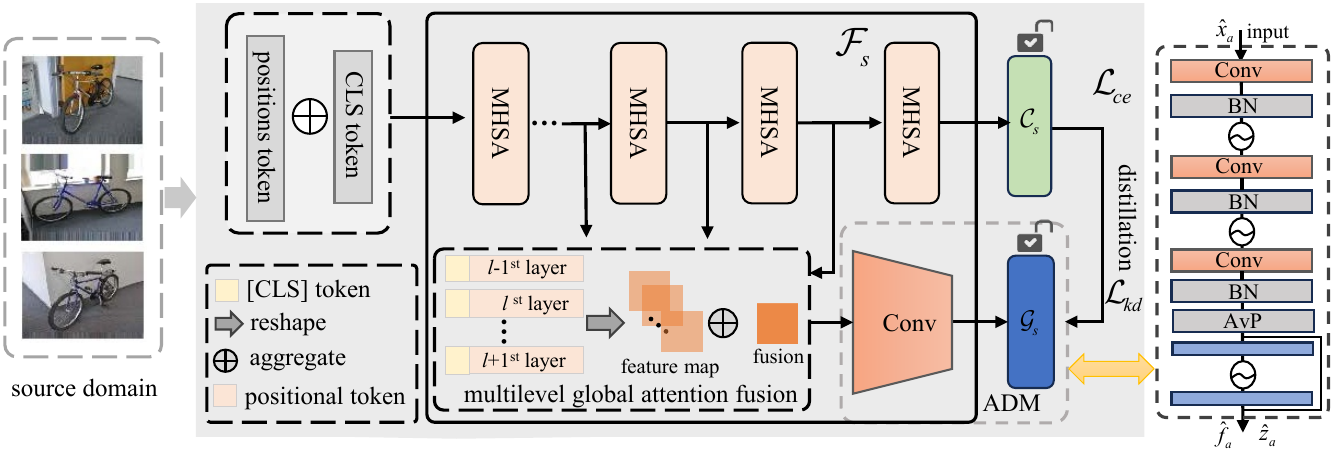}
   \caption{The overall workflow of the proposed CADTrans. In the initial stage, CADTrans undergoes training and distillation within the source domain, where the attention features from each layer are aggregated through EMA to generate global attention. The ADM block $\mathcal{G}_{s}$ is the trainable module by distillation of the output of classifier $\mathcal{C}_{s}$.}
   \label{fig:framework}
   \vspace{-4mm}
\end{figure*}

In this paper, we propose a {\textbf{C}onsistent} \textbf{A}ssistant \textbf{D}omains \textbf{Trans}former for SFDA, abbreviated as \textbf{CADTrans}, which solves the issue by constructing invariable feature representations on domain consistency.
The reason for adopting the vision transformer (ViT) to study related problems is primarily based on the following considerations:
First, compared to the CNN-based model, the inner feature dimensions of each layer in the ViT backbone are consistent, enabling the relatively low cost of constructing the assistant domain.
Second, we argue that self-attention features can capture discriminative features, which surpass general semantic features in terms of representation power.
Besides, since the long-distance dependence of features is alleviated, the current ViT model has achieved immense success in many fields~\cite{dosovitskiy2020image,carion2020end,zheng2021rethinking}, while related research has been relatively less explored in the field of SFDA.
Specifically, CADTrans first acquires the multilevel global attention from the intermediate layers via exponential moving average~(EMA), and then introduces 
a plug-and-play assistant domain module~(ADM) for the ViT backbone to obtain diverse representations from the aggregated global attentions, which can be used to construct an assistant domain, as depicted in Fig.~\ref{fig:framework}.
We train the CADTrans in the source stage~(i.e., first stage) by supervised learning and self-distillation. Then we initialize the target model with the parameters of the source domain, fixing the ADM and the target domain module to conduct domain adaptation.
In light of the assistant domain and target domain of domain adaptation stage~(i.e., second stage), we distinguish invariable feature representation by employing multiple consistency strategies, i.e., dynamic consistent labels and consistent neighbors, which can divide the whole target samples into source-like easy samples and target-specific hard samples.
Finally, we align the hard samples to the easy samples by conditional multi-kernel max mean discrepancy~(CMK-MMD), which guides the hard samples to align corresponding easy samples.

To verify the effectiveness, we conduct extensive experiments on various scale benchmarks, i.e., Office-31, Office-Home, VISDA-C, and {DomainNet-126}. Eventually, our approach can outperform other source-free approaches and achieve superior performance among multiple domain adaptation benchmarks.
To sum up, our key contributions are as follows:
\begin{itemize}
  \item {We construct an efficient transformer variant with an assistant domain to obtain diverse representations from intermediate layers to assist construct assistant domain, as well as effectively mitigating the lack of inductive bias in ViT-based backbone models.}
    \item {We first address the issue of SFDA from a new perspective by domain consistency and propose exploiting multi-consistent strategies to distinguish invariable feature representations. Although this method introduces some complexity,} it effectively divide the whole target samples into source-like easy samples and target-specific hard samples.
  \item {To eliminate the domain shift and further improve the alignment effect of the domain adaptation, we align the hard to the easy samples by CMK-MMD, which guides the hard sample to align the corresponding easy samples. }
  \item {Extensive comparative experiments are conducted across various benchmarks to rigorously validate the effectiveness of our proposed methodologies. }
\end{itemize} 

\section{Related Work}
\textbf{Domain adaptation.} 
DA tackles the issue of domain shift by aligning the feature distribution in the common project feature space, which has been successfully applied in various cross-domain tasks. Currently, DA approaches mainly adopt the two paradigms, moment matching~\cite{long2015learning,sun2016return,flamary2016optimal,kang2019contrastive} and adversarial learning~\cite{ganin2016domain,tzeng2017adversarial,long2018conditional}. 
Moment matching aims to learn a domain-invariant representation by minimizing the domain discrepancy, such discrepancy metric includes maximum mean discrepancy (MMD)~\cite{long2015learning} as well as contrastive domain discrepancy~\cite{kang2019contrastive}, and the Wasserstein metric~\cite{flamary2016optimal} etc. 
e.g., MMD is a non-parametric kernel function embedded into reproducing kernel hilbert space (RKHS) to measure the difference between two probability distributions. 
Adversarial learning exploits the generator to construct domain-invariant representations, e.g., DANN~\cite{ganin2016domain} formulates domain adaptation as an adversarial two-player game. CDAN~\cite{long2018conditional} is conditioned on several sources of information.
However, these methods often require access to the source domain, which increases the risk of privacy information leakage. 
Consequently, to overcome this challenge, we tackle this issue from the aspect of SFDA.

\textbf{Source free domain adaptation.} 
SFDA is the sub-field of UDA~\cite{pan2009survey,ganin2015unsupervised,long2015learning}, which aims to adapt to a target domain without accessing the source domain directly, due to data privacy. 
Some prospective works have proposed to alleviate the issue from different aspects~\cite{liang2020we,li2020model,xia2021adaptive,yang2021generalized}.
Among them, adversarial-based approaches include 3C-GAN~\cite{li2020model}, which develops a collaborative class-conditional generative adversarial network to produce target-style training samples, and A$^{2}$Net~\cite{xia2021adaptive}, which adopts a soft-adversarial manner to adaptively distinguish source-similar target samples from source-dissimilar ones.
While pseudo-label evaluation approaches include 
SHOT~\cite{liang2020we}, which evaluates the pseudo-labels by constructing the centroid of coarse-grained clustering features, then aligns the source domain with the target domain by minimizing the information maximization loss, and G-SFDA~\cite{yang2021generalized}, which evaluates the category consistency by the nearest neighborhood features. 
{Recently, DaC~\cite{zhang2022divide} proposes to divide the target data into source-like and target-specific samples, and contrast them by contrast learning. While DSiT~\cite{sanyal2023domain} propose to induce domain-specificity for SFDA.
Different from DaC, we divide the easy and hard samples by consistent strategies of assistant domain. 
Opposite to DSiT, our approach aims to reconstruct the invariable feature representations for disentangling and aligning easy and hard samples. }

\textbf{Vision transformers for DA.} 
Transformer~\cite{vaswani2017attention} was first developed in the NLP field to solve the long-range dependence. 
Due to the excellent performance, the variants of ViT~\cite{dosovitskiy2020image} are continually introduced into the visual field and achieved great success in image classification~\cite{dosovitskiy2020image}, 
object detection~\cite{carion2020end}, and semantic segmentation~\cite{zheng2021rethinking}, etc. 
Recently, the application of ViT is observed in the field of  DA~\cite{yang2023tvt, xu2021cdtrans, sun2022safe}. e.g., CDTrans~\cite{xu2021cdtrans} proposes cross-attention for domain alignment,  TvT~\cite{yang2023tvt} exploit adversarial training while SSRT~\cite{sun2022safe} uses a self-training mechanism for DA task.  
However, the application in SFDA is still relatively scarce. 
Current exploration, TransDA~\cite{yang2023self} firstly introduces a self-attention layer for a ResNet backbone to solve SFDA, which is not a pure ViT-based solution. DSiT~\cite{sanyal2023domain} try to exploit the ViT as the backbone to enhance domain-specificity.
Differing from the traditional CNN-based DA models and transformer-based model~(i.e., TransDA, and DSiT), we design and exploit the new variant of ViT with ADM as the backbone, which obtains diversified representations from the coincident assistant domains to effectively distinguish between easy and hard samples.

{\textbf{Consistency-based strategies for UDA.}
Consistency-based strategies are extensively investigated studied in the semi-supervised learning tasks~\cite{luo2022domain, tang2023consistency, liu2024domain, lu2024consistency, liu2025consistency}, and the key point is similar inputs should share consistent predictions, which can enhance the network's feature representation. CRMA~\cite{luo2022domain} propose intra-domain consistency and inter-domain consistency to achieve alignment of among multi-source DA setting. Differently, CODA~\cite{liu2025consistency}, in multi-source SFDA setting, mainly leverages the label consistency criterion to regularize the soft labels of weakly- and strongly-augmented target samples from each pair of source models, allowing them to supervise each other.
Another CRMA~\cite{lu2024consistency} mainly propose consistency regularization-based mutual alignment in SFDA, which utilizes the augmented version of each target sample to achieve alignment. 
CODA~\cite{liu2025consistency} and CRMA~\cite{lu2024consistency} are all only exploit consistency regularization to achieve alignment from augmented target samples, while our approaches focus primarily on the distinction of features. CRMA~\cite{luo2022domain} solve the alignment of fine-grained sudomain in multi-source DA setting, whereas the approach of our method is more challenging in SFDA scenarios in comparison and thus requires the help of assistant domains.  
}

\section{Methods}
 \subsection{Preliminaries}
 In this paper, we address the unsupervised SFDA task only with a pre-trained source model and without accessing any source data.
 We denote the labeled source domain data with $n_{s}$ samples as $\mathcal{D}_{s} = { \left \{ (x_{s}^{i}, y_{s}^{i}) \right \} }^{n_{s}}_{i=1}$ where $x^{i}_{s} \in \mathcal{X}_{s}, y^{i}_{s} \in \mathcal{Y}_{s}$, and unlabeled target domain data with $n_{t}$ samples as $\mathcal{D}_{t} = { \left \{ x_{t}^{i}  \right \} }^{n_{t}}_{i=1}$ where $ x^{i}_{t} \in \mathcal{X}_{t} $. The goal of SFDA is to train a classifier of $\mathcal{C}  \circ \mathcal{F} $ with the given source data to predict the labels $\left \{  y^{i}_{t} \right \}^{n_{t}}_{i=1}  $ of unlabeled data in the target domain, where $\mathcal{F} (\cdot; \theta_{\mathcal{F} }): x^{i}_{t} \to f^{i}_{t} $ denotes the backbone model of features extractor, where $f_{t}$ denotes the feature representation. $\mathcal{C} (\cdot; \theta_{\mathcal{C}}):  f^{i}_{t} \to z^{i}_{t} $ denotes the linear classifier, and $z^{i}_{t}$ represents the distribution of logits, as shown in Fig.~\ref{fig:framework}. Besides, we introduce an ADM block as $\mathcal{G} (\cdot; \theta_{\mathcal{G}}): \hat{x}^{i}_{a} \to \hat{f}^{i}_{a}$ and $\hat{z}^{i}_{a}$, where $\hat{x}_{a}^{i}$ denotes global attention feature, $\hat{f}^{i}_{a}$ and $\hat{z}^{i}_{a}$ denotes the feature representations and the distribution logits extracted by the ADM.%

\noindent \textbf{Multi-head self-attention.} Self-attention mechanism is a core component of ViT, which captures the long-distance dependence by MHSA.
For self-attention, the layer-normalized patches are mapped by three learnable linear projectors into vectors, i.e. queries $\mathbf{Q} \in \mathcal{R}^{(\mathrm{N+1}) \times \mathrm{D}}$, keys $\mathbf{K} \in \mathcal{R}^{(\mathrm{N+1}) \times \mathrm{D}}$ and values $\mathbf{V} \in \mathcal{R}^{(\mathrm{N+1})  \times \mathrm{D}}$. $\mathrm{N}$ donates the length of patch sequence, $\mathrm{D}$ indicates the dimensions of $\mathbf{\mathrm{Q}}$ and $\mathbf{\mathrm{K}}$.
Then the queries and keys are matched by the inner product to have an ${(\mathrm{N+1}) \times (\mathrm{N+1})}$ matrix, which denotes the semantic relevance of the query-key pairs in the corresponding position and applies a softmax function to obtain the weights on the values, which is given by:
 \begin{align}
     \mathrm {SA}(\mathbf{Q}, \mathbf{K},\mathbf{V})&=softmax(\frac{\mathbf{QK}^{\mathrm{T}}}{\sqrt{D} } )\cdot \mathbf{V} \,,  \\   \notag
     \mathrm {MHSA}(\mathbf{Q}, \mathbf{K},\mathbf{V}) & = \mathrm{Concat}(\mathrm{SA}_{1},...,\mathrm{SA}_{k})   \,,
 \end{align}
 $\mathrm{Concat}$ here represents a concatenation operation. In our solution, we learn the representations from the multilevel global attention fusion as detailed in the following. 

\subsection{Assistant domain construction and training}
{As mentioned above, owing to the inability to access the source domain data, compared to DA tasks~\cite{long2015learning}, the main dilemma of SFDA is the inability to obtain feature representations of the source domain directly, which makes it difficult to align the source and target domain by invariant  feature representations.
Therefore, existing SFDA methods~\cite{liang2020we,yang2021generalized,ding2022source} commonly exploit feature distribution of target domain to evaluate the source distribution or evaluate the pseudo-labels of target by clustering of features.
}
{Unlike current SFDA approaches, to obtain the diverse representations, we pay attention to the intermediate layers, which are widely exploited in current research~\cite{dosovitskiy2020image,chen2023cf,liang2022not} yet significantly overlooked in the SFDA field. }
We exploit the ViT models as our basic backbone.
Compared with the CNN models, ViT models excel in addressing the long-distance dependence issue of features, however, missing a certain inductive bias makes them require a large dataset~(e.g., ImageNet ILSVRC 2012 dataset~\cite{deng2009imagenet} or JFT-300M dataset~\cite{sun2017revisiting}) to compensate for this.
Therefore, to address the shortcoming of inductive bias and achieve diverse representations, we construct and introduce an extra CNN-based assistant domain module for CADTrans, which allows us to exploit the feature representations from the intermediate MHSA layers~\cite{dosovitskiy2020image,chen2023cf,liang2022not}.
Commonly, ViT model exploits the class token~(i.e., $ \mathrm{[CLS]}$ token) to learn the global positional information of all patches and utilize it for the final classification of objects.
However, some research~\cite{liang2022not,chen2023cf} have proven that not all layers of attention can focus on discriminative objects.
If directly achieving the feature of each layer will lead to a weak representation. 
Therefore, we take the features of the multilevel MHSA into account and aim to learn the feature representations from global self-attention. 
Specifically, we aggregate the attentional features of each layer by exponential moving average~(EMA) to achieve the global attention as $ \hat{x}_{a}=\lambda \cdot \hat{f}^{a}_{l} + (1-\lambda) \cdot \hat{f}^{a}_{l-1}$, where $\lambda=0.99$ and $l$ denotes the current layer of MHSA. We follow the suggestion of~\cite{liang2022not} and set $l > 4$-th layer to ensure the acquisition of effective attention features. Then, our CADTrans extracts the feature of the global attention by the assistant module as 
{
\begin{align}
  \begin{cases} 
  \hat{e}_{1} &= \mathrm{\Phi}(\mathrm{\Psi }(\mathcal{A}^{3 \times 3}(\hat{x}_{a}))) \,, \\ 
  \hat{e}_{2} &= \mathrm{\Phi}(\mathrm{\Psi}(\mathcal{A}^{3 \times 3}(\hat{e}_{1})))  \,, \\
  \hat{f}_{a} &= \mathrm{\varTheta }(\mathrm{\Psi}(\mathcal{A}^{3 \times 3}(\hat{e}_{2}))) \,, \\
  \hat{z}_{a} &= \mathrm{\varUpsilon }(\mathrm{\Phi}(\mathrm{\varUpsilon }(\hat{f}_{a}))) \,,
  \end{cases}   
  \label{ep:adm}
\end{align}
where $\hat{e}_{1}$ and $\hat{e}_{2}$ indicate the internal latent vectors of the ADM, $\mathcal{A}^{3 \times 3}$ denotes the convolution with a kernel size of $3 \times 3$, $\Psi$, $\Phi$, $\varTheta$, and $\varUpsilon$ denote the BatchNormal layer, Rectified Linear unit, Average Pool layer, and Linear layer, respectively. }
{The ADM block of the CADTrans backbone is trained during the initial stage~(i.e., first stage) of the source domain and remains fixed for the adaptation to the target domain. This process allows us to obtain diverse feature representations, denoted as $\hat{f}_{a}$, and distribution logits, represented as $\hat{z}_{a}$, which assist in constructing the assistant domain.}

\begin{figure*}[!ht]
  \centering
   \includegraphics[width=\linewidth]{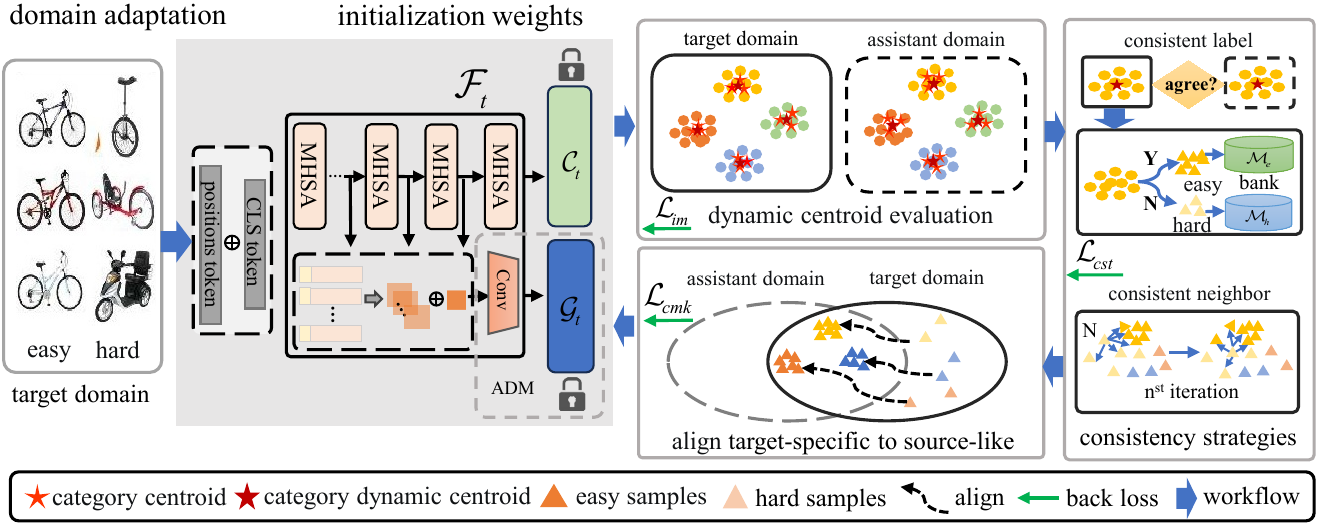}
   \vspace{-2mm}
   \caption{ The second stage involves target adaptation. We first evaluate the pseudo-labels from both target domain and assistant domain to distinguish easy samples~(dark color) and hard samples~(light color) by consistency strategies. Then, we store the easy and hard samples in the memory bank respectively and reassess the hard samples by consistent neighbors. Finally, we exploit CMK-MMD to align the hard samples and easy samples.
   }
   \label{fig:framework_2}
   \vspace{-4mm}
\end{figure*}

\textbf{First stage training and self-distillation.} 
In the first stage of our framework, we train the model in the source domain with ground truth by standard cross-entropy loss and aim to obtain the optimal ADM parameters $\theta _{\mathcal{G}}^{*}$ by minimizing the discrepancy between ADM $\mathcal{G}$ and source classifier $\mathcal{C}$ as
\begin{equation}
    \theta _{\mathcal{G}}^{*}= \mathop{\arg\min}\limits_{\theta_{\mathcal{G}}^{*}}\mathbb{E}_{x}[\mathcal{D}(\mathcal{G}( \hat{x}_{a};\theta_{\mathcal{G}}^{*}), \mathcal{C}( f_{s} ;\theta_{\mathcal{C}}) \circ \mathcal{F}( x_{s} ;\theta_{\mathcal{F}}))] \,,
    \label{ep:miniadm}
\end{equation}
where the $\theta_{\mathcal{G}}^{*}$ denotes the optimized weight of the ADM block, which is trainable in the source domain.
To realize this, we exploit the self-distillation strategy to optimize the weight $\theta_{\mathcal{G}}^{*}$ of ADM, which distills the knowledge from the feature $f^{i}_{s}$ and logits $z^{i}_{s}$ of classifier $\mathcal{C}$. Therefore, in the source domain, we train the model by the formula as follows: 
\begin{align}
  \mathcal{L}^{s}_{kd} = \frac{1}{n_{s}}\sum_{i=1}^{n_{s}} \left( \|\hat{f}^{i}_{a,s} - f^{i}_{s} \|_{2} + \hat{z}^{i}_{a,s} \cdot \mathbf{log} \frac{\hat{z}^{i}_{a,s}}{z^{i}_{s}} \right)  \,,
  \label{eq:kd_mse}
\end{align}
where $ \hat{f}^{i}_{a,s}$ and $\hat{z}^{i}_{a,s}$ denote the feature and logits mapped by ADM in the source domain. According to our experiments in Tab~\ref{tab:SD_ADM}, CADTrans can significantly improve the performance of the source domain. {This fully demonstrates the ability of CADTrans to mitigate the lack of inductive bias, then we apply it in the second stage for target domain adaptation, as depicted in Fig.~\ref{fig:framework_2}.}

\subsection{Domain consistency strategies}
\textbf{Dynamic centroid evaluation.} 
In the domain adaptation progress, we first exploit CADTrans to extract the feature representations $f_{t}$ and $\hat{f}_{a,t}$ with corresponding logits distribution $z_{t}$ and $\hat{z}_{a, t}$ from unlabeled dataset $\mathcal{X}_{t}$. Then, we freeze the source classifier and ADM, i.e., $\mathcal{C}$ and $\mathcal{G}$. Recent studies have shown that if the pseudo-labels of target samples are calculated directly from centroid evaluation proposed by ~\cite{liang2020we}, it was easily disturbed by noise~\cite{zhang2022divide}. Thus, differing from the traditional approaches~\cite{liang2020we,liang2021source,zhang2022divide}, we exploit the dynamic strategy to calculate the initial centroid of each category by 
\begin{align}
  c^{j}_{k} &= \frac{\sum_{x_{t} \in \mathcal{X}_{t}} exp(z^{j}_{t}) f^{j}_{t}}{\sum_{x_{t} \in \mathcal{X}_{t}}exp(z^{j}_{t})} \,, \\   \notag
  c_{k}^{j} & = \lambda c_{k}^{j-1} + (1-\lambda)c_{k}^{j}  \,,
  \label{eq:dy_label}
\end{align}
where $c^{j}_{k}$ is the $k$-th category centroid of the $j$-steps explorations, and $exp(\cdot)$ is the exponential function, $\lambda$ is the coefficient of movement. Then we estimate the label of each sample by its nearest initial centroid as $\hat{\mathcal{Y}}_{t} =\{\hat{y}^{i}_{t} | \hat{y}^{i}_{t} = \arg \min_{k} D(f^{i}_{t}, c^{j}_{k}) \} $, where the $D(\cdot,\cdot)$ is the cosine similarity function to measure the distance between features and centroids. While the $k$-th centroid is further modified by 
$c^{j}_{k} = \frac{\sum_{x_{t} \in \mathcal{X}_{t}} \mathds{1} (\hat{y}_{t}^{j} = k) f^{j}_{t}}{\sum_{x_{t} \in \mathcal{X}_{t}} \mathds{1} (\hat{y}_{t}^{j} = k)}$, 
where $\mathds{1}$ is the indicator function. The pseudo-labels are continuously updated in $j$-step iterations. 

\textbf{Consistent labels evaluate easy samples.} 
With the pseudo-label evaluation strategy, we can obtain two individual pseudo-label sets as $ \{\hat{\mathcal{Y}}_{g}, \hat{\mathcal{Y}}_{c} \}$, where the $\hat{\mathcal{Y}}_{g}$ and $\hat{\mathcal{Y}}_{c}$ are the label lists obtained from the ADM block $\mathcal{G}$ and source classifier $\mathcal{C}$, respectively. 
These two pseudo-label sets come from different dimensions of the feature representation, 
which means that the features mapped by consistent labels can be treated as invariant features. 
Therefore, to define the categories of invariant features, we use the consistent pseudo-labels in the two spaces as the final pseudo-labels of the target samples by $\mathcal{\hat{Y}}_{e} = \mathds{1} \cdot \{\mathcal{\hat{Y}}_{g}, \mathcal{\hat{Y}}_{c} \}$, where $\mathds{1} \in \{0, 1\}$ is an indicator function that returns 1 if $\hat{y}^{i}_{g} = \hat{y}^{i}_{c}$ evaluates as true. While the rest inconsistent set can be denote as $\hat{\mathcal{Y}}_{h} = \mathcal{\hat{Y}}_{c} \setminus  \mathcal{\hat{Y}}_{e}$, where the $\mathcal{\hat{Y}}_{h}$ indicates the inconsistent hard labels. Then, we store the corresponding feature representations by the evaluated labels to easy sample bank $\mathcal{M}_{e}$ and hard sample bank $\mathcal{M}_{h}$. 
We argue that the sample discriminated by both domains is an easy sample that is more related to the source, while the rest samples are hard to discriminate or easily confused samples. Besides, our divided approach relies on the consistency of the two domains, which makes it different from the threshold method employed by DaC~\cite{zhang2022divide}.

\textbf{Consistent neighbors reassess hard samples.} 
With the above consistent strategy, confident pseudo-labels are obtained for easy samples. However, hard samples are more prone to be mislabeled, potentially causing domain shift, so further strategies are needed to reassess their labels.
{As shown in Fig~\ref{fig:overview}, the hard samples in the target domain differ significantly from those in the source domain due to background interference and other factors. However, they share a consistent distribution with the easy samples in the target domain, which means they still retain some similarities.}
To better identify inconsistent samples, we propose to compare the labeled easy samples with susceptible hard samples, and help the hard samples reassess the labels. 
Therefore, we exploit the rating matrix to measure the similarity scores of the space to vote the consistent labels.
Concretely, all target feature representations $f_{t}$ have been divided into easy bank feature representations $\mathcal{M}_{e}$ and hard bank feature representations $\mathcal{M}_{h}$ by marked consistent labels. Then, we calculate the rating matrix of the divided feature representations in memory bank by $ \mathcal{S}_{n \times m} = \{ \mathcal{M}_{h} \cdot \mathcal{M}_{e}^{\top } \}$, where the $\mathcal{S}_{n\times m}$ indicates the similarity score of the features while $n$ and $m$ are count of corresponding feature representations s.t. $ n + m = n_{t}$. 
For each inconsistent label $\hat{y}_{h} \in \hat{\mathcal{Y}}_{h}$, we exploit consistent label of its sampled $k$-nearest neighbors as $\tilde{y}^{k}_{e} = \mathcal{\hat{Y}}_{e}^{\arg \max_{k} \mathcal{S}} $, where $k$ is the index of the nearest neighbors and s.t. $k < m$. The reassessed hard labels are obtained by $\hat{y}_{h} = \hat{y}^{k}_{e}$. Then we combine the reconstructed pseudo-labels with the consistent labels to obtain the final pseudo-labels $ \tilde{y}_{t} \in \tilde{\mathcal{Y}}_{t}$. Different from directly local neighbors of \cite{yang2021generalized,wang2022exploring}, our neighbors are calculated from the divided easy and hard samples, where labeled easy samples provide a more definitive assessment. 

\textbf{Consistent self-supervised training.}
With the above strategies, we can obtain high-confidence pseudo-labels, which will benefit the training of self-supervised processes. Therefore, we utilize the pseudo-labels to supervise the consistent training and final alignment, the self-supervised training loss is combined with two items as follows:
  \begin{equation}
  \mathcal{L}^{t}_{cst} = \frac{1}{n_{b}} \left(  \sum_{i=1}^{n_{b}} \mathrm{CE}(\hat{z}^{i}_{t},  \tilde{y}^{i}_{t}) +  \sum_{i=1}^{n_{b}} \mathrm{CE}(z^{i}_{t},  \tilde{y}^{i}_{t} ) \right) \,,
  \label{eq:cls_loss}  
  \vspace{-2mm}  
  \end{equation}  
where $n_{b}$ is batch-size of each mini-batch. The first item of $CE(\cdot,\cdot)$ is used to optimize the model from assistant domain of $\mathcal{G}$, while the second item is used to optimize the model from target domain of $\mathcal{C}$.
By $\mathcal{L}_{cst}$, classificatory samples can be aligned in different spaces, thereby enhancing their discriminative consistency across various dimensions.

\subsection{Align hard samples to easy samples}
To solve the domain shift and encourage the feature representation to be invariant across different domains, traditional methods solve the challenge of DA or SFDA by aligning the discrepancies between the two domains~\cite{long2015learning,kang2019contrastive,flamary2016optimal}. 
Back to the principle depicted in Fig.~\ref{fig:overview}, easy samples are closer to the source domain, while hard samples are specific to the target domain. 
Therefore, aligning the source domain with the target domain can be achieved by the alignment of hard samples and easy samples, which changes the SFDA problem to the traditional DA problem. 
Then, we propose an improved discrepancy for measuring the
distribution difference based on traditional MMD~\cite{long2015learning}. %

\textbf{Definition 1.}~\textit{Assume that easy samples as the source domain $\mathcal{X}_{e} \sim \tilde{\mathcal{X}}_{s}$ and hard samples as the new target domain, $\mathcal{X}_{h} \sim \tilde{\mathcal{X}}_{t}$, where ${\tilde{x}_{s}^{i}} \in \tilde{\mathcal{X}}_{s}$ and ${\tilde{x}_{t}^{j}} \in \tilde{\mathcal{X}}_{t}$ are $i.i.d.$,  the corresponding representations $\tilde{f}^{i}_{s}$ and $\tilde{f}^{j}_{t}$ are sampled from the $\mathcal{M}_{e}$ and $\mathcal{M}_{h}$, where $i \in n$, $j \in m$, and $n+m=n_{t}$. 
The MMD between two distributions and their mean embedding in the reproducing kernel hilbert space~(RKHS) is defined as} 
\begin{equation}
D_{\mathcal{H}}(\tilde{f}_{s}, \tilde{f}_{t}) = \sup _{\phi \sim \mathcal{H}} ( \mathbb{E}_{\tilde{\mathcal{X}}_{s}}[\phi(\mathcal{F}_{t}(\tilde{\mathcal{X}}_{s} )) ]-\mathbb{E}_{\tilde{\mathcal{X}}_{t}}[\phi(\mathcal{F}_{t}(\tilde{\mathcal{X}}_{t}))])_{\mathcal{H}}, 
\end{equation}
where $\mathcal{H}$ is the hypothesis space while the $\phi(\cdot)$ indicates the feature map.
Having assessed the labels of easy and hard samples, we aim to enhance domain adaptation further. To achieve this, we incorporate category information into MMD to produce a new variant, resulting in a new variant named CMK-MMD.
This approach adaptively guides different features, reducing intra-class variation while augmenting inter-class variation.

\textbf{Definition 2.}\textit{
For any two domains of feature distributions $\tilde{f}_{s} $ and $\tilde{f}_{t}$ with corresponding labels $\tilde{y}_{s}$ and $\tilde{y}_{t}$, we re-define the CMK-MMD between them as follows: }
  \begin{equation}
      D_{K}^{2}(\zeta^{i}_{s}, \zeta^{j}_{t}) := \left\|\mathbb{E}_{\tilde{f}^{i}_{s}} \left[ \phi\left( \tilde{f}^{i}_{s}  | \tilde{y}^{i}_{s}\right) \right]-\mathbb{E}_{\tilde{f}^{j}_{t}}\left[\phi\left(\tilde{f}^{j}_{t}  | \tilde{y}^{j}_{t}\right)\right]\right\|_{\mathcal{H}}^{2} \\,
  \end{equation}
\noindent where $D^{2}_{k}(\zeta^{i}_{s},\zeta^{j}_{t}) = 0$ iff $\tilde{f}^{i}_{s}=\tilde{f}^{j}_{t} $ s.t. $ \tilde{y}^{j}_{s} = \tilde{y}^{j}_{t}$ while the kernel associated with the feature maps $\phi$ as $K(\zeta_{s}, \zeta_{t}) =  \langle   \phi(\tilde{f}_{s} | \tilde{y}_{s}), \phi(\tilde{f}_{t} | \tilde{y}_{t}) \rangle $. 
Based on this, multi-kernels can defined as  
 $\mathcal{K}:=\left\{K=\sum_{u=1}^{v} \gamma_{u} K_{u}: \gamma_{u} \geq 0, \forall u \right\} $, 
where $v$ is the number of kernels, $\gamma_{u}$ is the weight coefficients used to constrain the kernel function, 
which can leverage the different kernels to satisfy CMK-MMD. 

\begin{algorithm}[!htbp]
  \caption{Pipeline of CADTrans on the source domain.}
  \label{alg:algorithm_1}
  \textbf{Input}: Source data $x^{s}_{i} \in \mathcal{X}^{s}_{i}$ with $ \mathcal{F}_{s}$, $ \mathcal{C}_{s} $ and  $\mathcal{G}_{s}$ and iteration $T_{s}$.\ \\
  \textbf{Output}: Trained model of source domain $ \mathcal{F}_{s}$, $ \mathcal{C}_{s} $ and  $\mathcal{G}_{s}$.\\
  \textbf{Initializtion}: Initialize $\mathcal{F}_{t}$ with pre-trained parameters.
 
  \begin{algorithmic}[1]
  \FOR{\textit{epoch = 1} to $T_{s}$}
  \STATE \textit{Step 1}. Aggregate the features of each layer to produce global feature maps $\hat{x}_{a}$ of by EMA.  
  \STATE \textit{Step 2}. Extract the feature representations $\hat{f}_{a,s}$ and logits $\hat{z}_{a,s}$  from multilevel global attention fusion by ADM block $\mathcal{G}_{s}$.
  \STATE \textit{Step 3}. Extract feature representations $f_{s}$ and logits $z_{s}$ $\mathcal{F}_{s}$ by source classifier $\mathcal{C}_{s}$.
  \STATE \textit{Step 4}. Train the source domain by supervised labels.
  \STATE \textit{Step 5}. Distill knowledge to optimize the ADM by  Eq.~\ref{eq:kd_mse}\\
  \ENDFOR
  \end{algorithmic}
\end{algorithm}

\begin{algorithm}[!ht]
  \caption{Pipeline of CADTrans on the target domain.}
  \label{alg:algorithm_2}
  \textbf{Input}: A pre-trained model $\mathcal{F}_{s}$, $\mathcal{C}_{s}$ and $\mathcal{G}_{s}$ on the source domain, Target domain $\mathcal{X}_{t}$, Hyper-parameters $\alpha$, $\beta$, iteration $T_{t}$. \\
  \textbf{Output}: Trained model for target domain $ \mathcal{F}_{t}$, $ \mathcal{C}_{t} $ and  $\mathcal{G}_{t}$.\\
  \textbf{Initializtion}: Initialize $\mathcal{F}_{t}$ with parameters pre-trained on the source domain, and freeze the classifier layer $\mathcal{C}_{t}$ and $\mathcal{G}_{t}$.
 
  \begin{algorithmic}[1]
  \FOR{\textit{epoch = 1} to $T_{t}$}
  \FOR{\textit{i = 1} to $n_{t}$}
  \STATE \textit{Step 1}. Initiate the memory bank with full representations.
  
  \STATE \textit{Step 2}.  Extract feature representations $\hat{f}_{a,t}$ and logits $\hat{z}_{a,t}$ by $\mathcal{G}_{t}$ to construct an assistant domain.
  \STATE \textit{Step 3}. Extract feature representations $f_{t}$ from output features and logits $z_{t}$ by $\mathcal{F}_{t}$ and $\mathcal{C}_{t}$.
  \STATE \textit{Step 4}. Dynamic centroid evaluation to determine pseudo-labels of both domains.
  \STATE \textit{Step 5}. Distinguish easy and hard samples by consistent strategies and store in the memory bank $\mathcal{M}_{e}$, $\mathcal{M}_{h}$, respectively.  
  \STATE \textit{Step 6}. Reassess hard samples by consistent neighbors.
  \ENDFOR
  \STATE \textit{Step 7}. Optimize the whole model by information maximization with loss function $\mathcal{L}_{im}$~( i.e., Eq.~\ref{eq:im_loss}).\\
  \STATE \textit{Step 8}. Train the target domain with pseudo-labels by consistent self-supervised training with $\mathcal{L}_{cst}$~(i.e., {Eq.~\ref{eq:cls_loss}}).
  \STATE \textit{Step 9}. Align the target-specific domain to source-like domain by $\mathcal{L}_{cmk}$~( i.e., Eq.~\ref{eq:cmk_loss}).\\
  \ENDFOR
  \end{algorithmic}
\end{algorithm}

\textbf{Definition 3.}\textit{
 Therefore, for any easy and hard feature representation $\tilde{f}^{i}_{s}, \tilde{f}^{j}_{t} $ with categories. CMK-MMD measures the discrepancy between representations of the easy and hard defined as:}
  \begin{align}
  \label{eq:cmk_loss}
  \mathcal{L}^{t}_{\mathrm{cmk}}  & =\frac{1}{n^{2}} \sum_{i=1}^{n} \sum_{j=1}^{n} \mathcal{K} \left(  \zeta^{i}_{s} , \zeta^{i}_{s} \right)  +\frac{1}{m^{2}} \sum_{i=1}^{m} \sum_{j=1}^{m} \mathcal{K} \left(  \zeta^{j}_{t} , \zeta^{j}_{t} \right) \\ \notag
  & -\frac{2}{n m} \sum_{i=1}^{n} \sum_{j=1}^{m} \mathcal{K} \left(  \zeta^{i}_{s} , \zeta^{j}_{t} \right) \,.
  \end{align}
\noindent\textbf{Remarks.} It is worth mentioning that recent  studies, such as SFDA-DE~\cite{ding2022source} and DaC~\cite{zhang2022divide}, also propose variants of MMD tailored to their specific settings. Nevertheless, SFDA-DE aligns the distributions with conditional MMD by the local feature while DaC aligns the global in a condition-free setting. Overall, our approach incorporates conditional information and aligns distributions on a global scale, effectively merging the benefits of both strategies. 
So far, despite being explored in several papers~\cite{long2015learning,flamary2016optimal,long2017deep,kang2019contrastive,zhu2020deep}, due to inaccessibility of source data, there has been no attempt to enhance the multiple feature representation in transformer-based architecture for SFDA with MMD, specifically from the perspective of distinguishing between easy and hard samples.

\newcolumntype{Y}{>{\centering\arraybackslash}X}
\begin{table*}[t]
  \centering
  \setlength{\tabcolsep}{8.7pt}
  \renewcommand{\arraystretch}{0.95}
  \caption{Accuracy~(\%) of different methods of unsupervised SFDA in Office-31 and VISDA-C datasets. The backbone of the Upper is ResNet-50 (RN-50) for Office-31 and ResNet-101 (RN-101) for VisDA. The results marked by `$\ast$'  come from our re-implementations. `SF' indicates `Source-free' and `SO' indicates `Source-only'. }
  \label{tab:ssda_o31}
  \resizebox{0.85\linewidth}{!}{
      \setlength{\extrarowheight}{1pt}
      \begin{tabular}{lcccccccccll}
        \Xhline{1.1pt}     
          \multirow{2}{30pt}{\centering Method} & \multirow{2}{*}{\centering SF} && \multicolumn{7}{c}{\textbf{Office-31}} && \textbf{VisDA}\\
          \cmidrule{4-10} \cmidrule{12-12}
          &&& A$\rightarrow$D & A$\rightarrow$W & D$\rightarrow$A & D$\rightarrow$W & W$\rightarrow$A & W$\rightarrow$D & Avg. && S$\rightarrow$R \\
          \midrule
          RN-50/101~(SO)~\cite{he2016deep} & $-$ & & 68.9 & 68.4 & 62.5 & 96.7 & 60.7 & 99.3 & 76.1 & & 52.4\\
          3C-GAN~\cite{li2020model} & $\checkmark $& & 92.7  & 93.7  & 75.3  & 98.5  & 77.8  & 99.8  & 89.6 & & 81.6\\
          SHOT~\cite{liang2020we}& $\checkmark $ & & 94.0  & 90.1  & 74.7  & 98.4  & 74.3  & 99.9  & 88.6 & & 82.9\\
          DIPE~\cite{wang2022exploring} & $\checkmark $ &  & 96.6 & 93.1 & 75.5 & 98.4  & 77.2 & 99.6 & 90.1 & & 83.1\\
          A$^2$Net~\cite{xia2021adaptive} & $\checkmark $ & & 94.5 & 94.0 & 76.7 & \textbf{99.2} & 76.1 & 100.0 & 90.1 & & 84.3\\
          NRC~\cite{yang2021exploiting} & $\checkmark $ & & 90.8 &  \textbf{96.0}  & 75.0 &  99.0 & 75.3 & 100.0 & 89.4 & & 85.9 \\
          SFDA-DE~\cite{ding2022source}  & $\checkmark $ & & 96.0 & 94.2 & 76.6 & 98.5 & 75.5 & 99.8 & 90.1 & & 86.5 \\
          SHOT++~\cite{liang2021source} & $\checkmark $ & & 94.3  & 90.4  & 76.2  & 98.7  & 75.8  & 99.9  & 89.2 & & 87.3 \\
          CRMA~\cite{lu2024consistency} & $\checkmark $ & & 94.4 &	92.5	& 75.9 &	99.0 & 76.4 & 100.0 &	89.7 & & 87.4 \\
          Improved SFDA~\cite{mitsuzumi2024understanding} & $\checkmark $ & & 	95.3 &	94.2 & 	98.3 &	99.9 &	76.4 & 	77.5 &	90.3 & & 88.4 \\
       CR-SFDA~\cite{tang2023consistency}  &  $\checkmark $ & &  -- &	-- & 	-- &	-- &	-- & 	-- &	-- & &  \textbf{88.7} \\
          \textbf{CADRN-50/101} &$\checkmark $ &   & \textbf{95.6}	& 94.7 &	\textbf{76.9} &	98.6 &	\textbf{77.7} &	\textbf{100.0} &	\textbf{90.6}&  & 87.6 \\
          \midrule
          ViT-B~(SO)~\cite{dosovitskiy2020image} &$ - $ & & 88.0 & 90.1 & 71.5 & 95.2 & 73.4 & 99.2 &	86.2  &  & 66.1 \\   
          DIPE (ViT-B)~\cite{wang2022exploring} &  $\checkmark $ & &	95.5 &	94.6 &	75.3	& 97.8	& 73.6	& 99.6	& 89.4	& & 82.8 \\
          TransDA~\cite{yang2023self} & $\checkmark $ &   & 97.2  & 95.0  & 73.7  & \textbf{99.3}   & 79.3  & 99.6  & 90.7 &  & 83.0 \\
          G-SFDA~(ViT-B)$^{\ast}$~\cite{yang2021generalized} & $\checkmark $ &  & 95.6 &	96.4 &	79.6 &	98.1 &	78.3 &	99.6 &	91.3  &  & 85.8 \\    
          Improved SFDA~(ViT-B)$^{\ast}$~\cite{kundu2022balancing} & $\checkmark $ &  & 96.6 &	95.4 &	78.4	& 98.9	& 78.9 &	\textbf{100.0}	& 91.4  &  & 86.3\\    
          SHOT++~(ViT-B)$^{\ast}$~\cite{liang2021source}  & $\checkmark $ & & 96.2 &	95.7 &	79.6 &	99.0 &	80.7 &	99.8 &	91.8   &   & 87.6 \\
          \textbf{CADTrans~(ViT-B)} &$\checkmark $ & &  \textbf{97.8}  & \textbf{97.6} & \textbf{81.7} & 99.3 &  \textbf{81.7}  & 99.8 & \textbf{93.0}  &  & \textbf{87.8} \\
          \midrule
          DeiT-B~(SO)~\cite{touvron2021training} &$ - $ &  & 88.8 &	88.9 &	70.4 &	96.6 &	74.0 &	99.0 &	86.3 &  & 61.7 \\ 
          DIPE~(DeiT-B)~\cite{wang2022exploring} & $\checkmark $ & & 95.5 & 98.5 & 77.1 & 94.8 & 77.5 &  100.0 & 90.5 &  & 82.8 \\
          Mixup~(DeiT-B)~\cite{kundu2022balancing} & $\checkmark $ &  & 96.1 & 98.6 & 80.1 & 95.4 & 80.2 & 100.0 & 91.7   &  &  86.3\\  
          G-SFDA~(DeiT-B)$^{\ast}$~\cite{yang2021generalized} & $\checkmark $ & & 96.2	& 96.4	& 82.7	& 98.0	& 78.4	& 99.4	& 91.9 &  & 86.6 \\
          SHOT++~(DeiT-B)$^{\ast}$~\cite{liang2021source}  & $\checkmark $ & & 96.4 &	97.2 &	80.8 &	98.7 &	79.8 &	99.6 &	92.1  &  & 87.9 \\
          Improved SFDA~(DeiT-B)$^{\ast}$~\cite{kundu2022balancing} & $\checkmark $ &  & 	96.9 &	98.4 &	82.3 &	98.1	& 81.4 &	100.0 &	92.9 & &	87.2 \\
          DSiT-B~\cite{sanyal2023domain} & $\checkmark $ &  &  97.2 &	\textbf{99.1} &	81.8 &	98.0 &	81.7 &	 \textbf{100.0} & 93.0 & & 87.6 \\  
          \textbf{CADTrans~(DeiT-B)} &$\checkmark $ & & \textbf{97.6} &	96.9	& \textbf{83.0} &	\textbf{99.1} &	\textbf{82.8} &	99.8 &	\textbf{93.2}  &   & \textbf{88.2}  \\ 
        \Xhline{1.1pt}          
      \end{tabular}       
      \label{tab:office-31}%
      }
    \vspace{-4mm}
\end{table*}

Besides, to make the target outputs individually certain and globally diverse, the common practice is to jointly perform feature learning, domain adaptation, and classifier learning by optimizing the information maximization~\cite{liang2020we} loss function as follows: 
\begin{align}
\label{eq:im_loss}
\mathcal{L}_{im} & = \mathcal{L}_{ent} (\mathcal{C} \circ \mathcal{F}; \mathcal{X}_{t}) +  \mathcal{L}_{div}(\mathcal{C} \circ \mathcal{F}; \mathcal{X}_{t}) \,, \\ \notag
& =  - \sum_{k=1}^{K} \delta(z_{t}^{i}) \log\delta(z_{t}^{i})  + \sum_{k=1}^{K} \hat{z}^{k}_{t} \log \hat{z}^{k}_{t},
\end{align}
where $\mathcal{L}_{div}$ denotes the diverse loss to avoid the same one-hot encoding, {$\hat{z}^{k}_{t}$ is the $k$-th element in a K-dimensional vector and obtained by  $\hat{z}_{t} = \frac{1}{n_{t}} \sum_{i=1}^{n_{t}} \delta(z_{t}^{i})$.} While $\mathcal{L}_{ent}$ means entropy loss to explore the possible samples, where $\delta(z^{i}_{t})$ is the softmax predicition of output logits $z^{i}_{t}$. 
With the above approaches, we can achieve the final total object loss:
\begin{align}
 \mathcal{L}_{total} = \mathcal{L}_{im} + \alpha \mathcal{L}_{cst} + \beta \mathcal{L}_{cmk} \,,
 \label{eq:total_loss}
\end{align}
where $\alpha$ and $\beta$ are the balance factor to control each item.

\section{Pipeline of the whole framework}
Finally, we list the algorithm implementation pipeline of the whole framework both on the source and target domain in the following \textbf{Algorithm}~\ref{alg:algorithm_1} and \textbf{Algorithm}~\ref{alg:algorithm_2}.

\subsection{Datasets and implementation details}
\label{sec:detail}
\textbf{Office-31}~\cite{saenko2010adapting} is a standard small-sized DA benchmark, which contains 4,652 images with 31 classes from three domains: Amazon (A), DSLR (D), and Webcam (W). 
\textbf{Office-Home}~\cite{venkateswara2017deep} is a medium-sized benchmark, which consists of 15,500 images with 65 classes from four domains: Artistic images (Ar), Clip Art (Cl), Product images (Pr), and Real-World images (Rw). 
\textbf{VisDA-C}~\cite{peng2017visda} is a challenging large-scale Synthetic-to-Real dataset that focuses on the 12-class object recognition task. The source domain contains 152 thousand synthetic images generated by rendering 3D models while the target domain has 55 thousand real object images sampled from Microsoft COCO. 
\textbf{DomainNet-126}~\cite{peng2019moment} 
is another large-scale dataset. As a subset of DomainNet containing 600k images of 345 classes from 6 domains of different image styles, which has 145k images from 126 classes, sampled from 4 domains, Clipart (C), Painting (P), Real (R), Sketch (S), as~\cite{saito2019semi} identify severe noisy labels in the dataset. 
Consistent with current mainstream benchmarks, our method is based on the ViT-B~(ViT-Base)/16 and Deit-B~(Deit-Base)/16, pre-trained on ImageNet-1k. The input size of the image in our experiments is $224 \times 224$, and each image is split into $16$ patches. Besides, the head account in each layer of our backbone is $12$.
{To further validate the effectiveness of our approaches, we reconstruct traditional CNN models by incorporating the ADM block into the ResNet architecture. For instance, we use ResNet-50 and ResNet-101 as our base models and modify them with our methods, resulting in the CADRN-50 and CADRN-101 architectures. }  %
The learning rate for Office-31 and Office-Home is set to 1e-2, while for VisDA-C and DomainNet is 1e-3. %
The training epoch is set to 100 and the whole process is optimized by stochastic gradient descent~(SGD) with a momentum of 0.9 and weight decay 1e-3. The batch size is 64 by default. The trade-off parameters $\alpha$, $\beta$ are set as 0.3, 0.1, respectively. All experiments are conducted with PyTorch on NVIDIA A800 GPUs. 

\subsection{Baselines and comparison methods}
We compare the proposed approach with the state-of-the-art methods mainly under the closed-set unsupervised SFDA.
For the closed-set DA, the compared methods include 
source-free methods of ResNet-50 and ResNet-101 backbones: 
SHOT~\cite{liang2020we}, 
3C-GAN~\cite{li2020model}, 
A$^2$Net~\cite{xia2021adaptive}, 
NRC~\cite{yang2021exploiting}, 
G-SFDA~\cite{yang2021generalized}, 
SHOT++~\cite{liang2021source}, 
DaC~\cite{zhang2022divide}, 
DIPE~\cite{wang2022exploring}, 
SFDA-DE~\cite{ding2022source}, 
CR-SFDA~\cite{tang2023consistency}, 
CRMA~\cite{lu2024consistency},
Improved SFDA~\cite{mitsuzumi2024understanding}
and some source-free methods of transformer backbone or re-implementations: 
TransDA~\cite{yang2023self}, 
DSiT-B~\cite{sanyal2023domain}, 
G-SFDA~(ViT-B)~\cite{yang2021generalized}, G-SFDA~(Deit-B), DIPE~(DeiT-B)~\cite{wang2022exploring}, Mixup~(DeiT-B)~\cite{kundu2022balancing} , SHOT++~(ViT-B)~\cite{liang2021source}, SHOT++~(Deit-B), Improved SFDA~(Deit-B)\cite{mitsuzumi2024understanding}.
Our ViT-B~\cite{dosovitskiy2020image} and Deit-B~\cite{touvron2021training} backbones are compared with the ResNet-50~\cite{he2016deep} backbone on small-sized and medium-sized datasets, i.e. Office-31 and Office-Home.
For the sake of fairness, we compare with ResNet-101 on the large-scale dataset, i.e. VisDA-C and DomainNet.
For each benchmark, we exploit the self-supervised training loss $\mathcal{L}_{im}$ as our baseline. `SO' represents exploiting the whole source model for target label prediction without domain adaptation.

\begin{table*}[!ht] 
  \centering
  \vspace{-3mm}
  \caption{Accuracy (\%) of different methods of unsupervised SFDA on the Office-Home dataset.}
  \resizebox{\linewidth}{!}{
    \begin{tabular}{lcccccccccccccc}
    \Xhline{1.1pt}
    \multicolumn{1}{c}{Method}& SF & Ar$\rightarrow$Cl &  Ar$\rightarrow$Pr &  Ar$\rightarrow$Rw  &  Cl$\rightarrow$Ar &  Cl$\rightarrow$Pr &  Cl$\rightarrow$Rw  & Pr$\rightarrow$Ar  & Pr$\rightarrow$Cl &  Pr$\rightarrow$Rw  & Rw$\rightarrow$Ar  & Rw$\rightarrow$Cl &  Rw$\rightarrow$Pr  & Avg. \\
    \midrule
    RN-50~(SO)~\cite{he2016deep}& $-$ & 34.9& 50.0& 58.0& 37.4& 41.9& 46.2& 38.5& 31.2& 60.4& 53.9& 41.2& 59.9& 46.1 \\
    SHOT~\cite{liang2020we}& $\checkmark$ & 57.1  & 78.1  & 81.5  & 68.0  & 78.2  & 78.1  & 67.4  & 54.9  & 82.2  & 73.3  & 58.8  & 84.3  & 71.8 \\
    G-SFDA~\cite{yang2021generalized}& $\checkmark$ &  57.9 & 78.6 & 81.0 & 66.7 & 77.2 & 77.2 & 65.6 & 56.0 & 82.2 & 72.0 & 57.8 & 83.4 & 71.3 \\
    NRC~\cite{yang2021exploiting}& $\checkmark$ &  57.7 & 80.3 & 82.0 & 68.1 & 79.8 & 78.6 & 65.3 & 56.4 & 83.0 & 71.0 & 58.6 & 85.6 & 72.2 \\
    DIPE~\cite{wang2022exploring} & $\checkmark$ &  56.5 & 79.2 & 80.7 & 70.1 & 79.8 & 78.8 & 67.9 & 55.1 & 83.5 & 74.1 & 59.3 & 84.8 & 72.5 \\
    A$^2$Net~\cite{xia2021adaptive} & $\checkmark$& 58.4 & 79.0 & 82.4 & 67.5 & 79.3 & 78.9 & 68.0 & 56.2 & 82.9 & 74.1 & 60.5 & 85.0 & 72.8 \\
    DaC~\cite{zhang2022divide} & $\checkmark$ & 59.1 & 79.5 & 81.2 & 69.3 & 78.9 & 79.2 & 67.4 & 56.4 & 82.4 & 74.0 & \textbf{61.4} & 84.4 & 72.8 \\
    CR-SFDA~\cite{tang2023consistency}  & $\checkmark$ & 58.6 & 80.2 & 82.9 & 69.8 & 78.6 & 79.0 & 67.8 & 55.7 & 82.3 & 73.6 & 60.1 & 84.9 & 72.8 \\
    SFDA-DE~\cite{ding2022source} & $\checkmark$ & 59.7 & 79.5 & 82.4 & 69.7 & 78.6 & 79.2 & 66.1 & 57.2 & 82.6 & 73.9 & 60.8 & 85.5 & 72.9 \\   
    SHOT++ ~\cite{liang2021source} & $\checkmark$ & 57.9 & 79.7 & 82.5 & 68.5 & 79.6 & 79.3 & 68.5 & 57.0 & 83.0 & 73.7 & 60.7 & 84.9 & 73.0 \\
    Improved SFDA~\cite{mitsuzumi2024understanding}	 & $\checkmark$ &  \textbf{60.7} & 78.9	& 82.0  &	69.9 &	79.5 &	79.7 &	67.1 &	\textbf{58.8} & 	82.3 & 	74.2	& 61.3 &	\textbf{86.4} &	73.4 \\
    CRMA~\cite{lu2024consistency} & $\checkmark$ & 59.5 & 79.9 & 82.0 & \textbf{72.1} & \textbf{82.1} & 80.1 & 68.8 & 57.6 & 84.1 & 73.9 & 61.0 & 86.0 & 73.9 \\
    \textbf{CADRN-50} & $\checkmark$ & 59.9 &	\textbf{81.3} &	\textbf{83.6}	& 70.5 &	79.4 &	\textbf{80.4} &	\textbf{69.4}	& 58.7	&  \textbf{84.9}	&  \textbf{74.6} &	60.6 &	85.8 &	\textbf{74.1} \\
    \midrule
    ViT-B~(SO)~\cite{dosovitskiy2020image} & $-$ & 51.4 &	81.1 &	85.4 &	73.6 &	82.2 & 83.0 & 73.6 & 50.8 & 87.2 & 78.2 & 50.1 & 86.4 & 73.6  \\
    DIPE~(ViT-B)$^{\ast}$~\cite{wang2022exploring} & $\checkmark$ & 64.1 &	81.3 &	83.6 &	68.9 &	80.8	& 79.2	& 68.6	& 61.7	& 85.3	& 75.8 &	63.1	& 88.3	& 75.1 \\
    G-SFDA~(ViT-B)$^{\ast}$~\cite{yang2021generalized} & $\checkmark$ & 63.9 & 81.0 &	84.5 &	73.3	&82.3 &	82.5	& 73.6 &	62.6	& 85.1	& 78.1	& 64.7 &	87.2	& 76.6  \\
    TransDA~\cite{yang2023self}& $\checkmark$ & 67.5  & 83.3  & 85.9  & 74.0  & 83.8  & 84.4  & 77.0  &   \textbf{68.0}  & 87.0  & 80.5  & 69.9  & 90.0  & 79.3 \\
    SHOT++~(ViT-B)$^{\ast}$~\cite{liang2021source} & $\checkmark$ & \textbf{68.7} &	87.2 &	 \textbf{88.4}	& 79.0 &	87.7 &	\textbf{93.1} &	78.8 &	59.0	&  \textbf{89.7} &	81.3	& 59.2	& 90.1	& 80.2   \\        
    Improved SFDA~(ViT-B)$^{\ast}$~\cite{mitsuzumi2024understanding}  & $\checkmark$ & 70.6	& \textbf{89.1} &	82.7 &	80.4	& 90.1	& 83.7 &	77.2	& 65.4 &	88.7 &	79.3 &	71.3 &	89.6 &	80.7 \\
    \textbf{CADTrans~(ViT-B)} & $\checkmark$  & 64.3 &	 87.8 &	86.3 &	 \textbf{82.5} &	 \textbf{91.8} &	 88.4 &	 \textbf{80.3} &	60.8 &	88.9 &	 \textbf{82.5} &	 \textbf{76.8} &	 \textbf{90.5} &	\textbf{81.7}  \\  
    \midrule
    DeiT-B~(SO)~\cite{touvron2021training}  & $-$ & 56.7 &	77.3 &	83.0 &	68.3 &	74.6 &	78.0 &	67.2 &	53.5 & 82.8 & 74.0 & 55.4 & 84.0 & 71.2  \\
    G-SFDA~(Deit-B)$^{\ast}$~\cite{yang2021generalized}& $\checkmark$ & 65.1 & 82.2 & 85.6 & 73.6 & 80.4 & 82.0 & 75.4 & 62.7 & 85.3 & 79.0 & 65.4 & 87.5 & 77.0  \\
    DIPE~(DeiT)$^{\ast}$~\cite{yang2021generalized} & $\checkmark$ & 66.0 & 80.6 & 85.6 & 77.1 & 83.5 & 83.4 & 75.3 & 63.3 & 85.1 & 81.6 & 67.7 & 89.6 & 78.2 \\
    Mixup~(DeiT)$^{\ast}$~\cite{kundu2022balancing} & $\checkmark$ & 65.3 & 82.1 & 86.5 & 77.3 & 81.7 & 82.4 & 77.1 & 65.7 & 84.6 & 81.2 & 70.1 & 88.3 & 78.5 \\
    Improved SFDA~(DeiT-B)$^{\ast}$~\cite{mitsuzumi2024understanding}  & $\checkmark$ & 	\textbf{71.3} &	\textbf{88.9}	& 83.8	& 79.4	& 89.3 &	83.2 &	76.8 &	63.0	& 88.6 &	79.1	& 67.1 &	88.7	& 79.9 \\
    DSiT-B~\cite{sanyal2023domain}  & $\checkmark$ & 69.2 & 83.5 & 87.3 &  80.7 & 86.1 & 86.2 & 77.9 &  \textbf{67.9} & 86.6 &  82.4 & 68.3 & 89.8 & 80.5 \\
    SHOT++~(Deit-B)$^{\ast}$~\cite{liang2021source}  & $\checkmark$ & 60.3 &	86.6 &	87.5 &	82.0 &	89.8 &	89.0 &	 \textbf{80.7} &	62.1 &	89.6 &	\textbf{82.6} &	 \textbf{77.1} &	90.5 &	81.5  \\
    \textbf{CADTrans~(DeiT-B)} & $\checkmark$ &  70.3 & 	 88.7 &	 \textbf{90.0} &	 \textbf{83.1} &	 \textbf{90.2} &	 \textbf{89.9} &	80.1 &	62.9 &	 \textbf{90.5} &	81.8 &	74.3 &	 \textbf{92.5} & \textbf{82.9}   \\
    \Xhline{1.1pt}
    \end{tabular}
  \label{tab:office-home}%
  \vspace{-4mm}
  }
\end{table*}%

\textbf{Results of Office-31.}
As can be seen from Table~\ref{tab:office-31}, 
For the ResNet-50 approaches in source-free settings, such as SHOT, 3C-GAN, A$^2$Net, DIPE, NRC, SHOT++, SFDA-DE, {CRMA}, and {Improved SFDA},  the best performance of these approaches can achieve around 90\%, which demonstrates SFDA can achieve good performance even in the scenario without source data. 
{On the office31 dataset, the effectiveness of our method in adapting to the three subdomains A$\rightarrow$D, D$\rightarrow$A, and  W$\rightarrow$A is significantly improved.  
Besides, it achieves an impressive performance of 90.6\% on average, demonstrating that the domain consistency strategy effectively differentiates between categories.}
Benchmarks in the middle of Table~\ref{tab:office-31} are some SOTA implementations based on the transformer, in which G-SFDA, Mixup, Improved SFDA, and SHOT++ are exploited ViT-B backbone to achieve a comparable performance with a score of around 91.0\%, indicating that the conventional pseudo-labeling strategy has reached its performance ceiling. This illustrates that architectural design plays a significant role in enhancing performance. Our CADTrans based on ViT-B can achieve the best performance of 93.0\%. 
Likewise, our CADTrans based on the DeiT-B backbone, at the bottom of Table~\ref{tab:office-31}, is a variant of DeiT-B models, which significantly outperforms previously published SOTA approaches, advancing the average accuracy from 93.2\%.
Furthermore, both methods have produced excellent results in various subfields, showcasing the superiority of our approach.

\begin{table*}[!htbp] 
  \caption{ Accuracy (\%) of different methods of unsupervised SFDA on {\bf DomainNet-126} dataset.   }
  \label{tab:fourdn}
  \resizebox{\linewidth}{!}{
  \centering
  \begin{tabular}{ l c c c c  c c c c c c c c c c }
      \Xhline{1.1pt}
      Method &{\bf S.F.}
      &C$\to$P &C$\to$R &C$\to$S
      &P$\to$C &P$\to$R &P$\to$S    
      &R$\to$C &R$\to$P &R$\to$S  
      &S$\to$C &S$\to$P &S$\to$R &Avg.\\
      \midrule
      RN-50 (SO)~\cite{he2016deep}     &--   & 48.8	& 62.7  & 50.5 &	59.4 & 76.0	 &	50.5 &	57.6 &	64.6	 &	50.4	&60.1 &	54.2 &	62.4 &	58.1 \\
      AdaCon~\cite{chen2022contrastive}&\checkmark  &60.8 &74.8 &55.9 &62.2 &78.3 &58.2 &63.1 &68.1 &55.6 &67.1 &66.0 &75.4 &65.4 \\
      PLUE~\cite{Litrico_2023}         &\checkmark  &59.8 &74.0 &56.0 &61.6 &78.5 &57.9 &61.6 &65.9 &53.8 &67.5 &64.3 &76.0 &64.7  \\
      TPDS~\cite{tang2024source}       &\checkmark  &62.9 &77.1 &59.8 &65.6 &79.0 &61.5 &66.4 &67.0 &58.2 &68.6 &64.3 &75.3 &67.1 \\
      NRC~\cite{yang2021nrc}           &\checkmark  &62.6 &77.1 &58.3 &62.9 &81.3 &60.7 &64.7 & \textbf{69.4} &58.7 &69.4 &65.8 &78.7 &67.5 \\
      SHOT~\cite{liang2020we}          &\checkmark  &63.5 &78.2 &59.5 &67.9 &81.3 &61.7 &67.7 &67.6 &57.8 &70.2 &64.0 &78.0 &68.1 \\
      CoWA~\cite{lee2022confidence}    &\checkmark  &64.6 &\textbf{80.6} &60.6 &66.2 &79.8 &60.8 &69.0 &67.2 &60.0 &69.0 &65.8 &\textbf{79.9} &68.6\\
      GKD~\cite{tang2021model}         &\checkmark  &61.4 &77.4 &60.3 &69.6 &  \textbf{81.4} &63.2 &68.3 &68.4 &59.5 &71.5 &65.2 &77.6 &68.7 
      \\
      Improved SFDA$^{\ast}$~\cite{mitsuzumi2024understanding} &\checkmark & 64.3	& 79.4	& 61.1 &	81.0 &	63.1 &	\textbf{67.9} &	68.1 &	59.9 &	\textbf{61.4}	&  72.6 &		65.9 &	79.3 &	68.7 \\  
      \textbf{CADRN-50}                & \checkmark &\textbf{65.3} &	80.2 &	 \textbf{61.6} &	 \textbf{70.3}	& 81.2 &	 64.2 &	 \textbf{69.2} &	68.8 &	60.7 &	\textbf{71.9} &	\textbf{66.7} & 	79.1 &	\textbf{69.9} \\
      \midrule
      ViT-B~(SO)~\cite{dosovitskiy2020image} &--&65.5 &	79.5 &	56.6 &	59.2 &	85.6 &	49.8 &	57.0 &	71.2 &	47.8 &	64.1 &	66.8 &	79.9 &	65.2 \\  
      DIPE (ViT-B)$^{\ast}$~\cite{wang2022exploring}& \checkmark & 64.9 &	81.7 &	65.3 &	71.8 &	82.5 &	65.4	& \textbf{68.9} &	69.0	& 59.7 &	70.6 &	65.3 &	77.2 & 70.2 \\
      TransDA~\cite{yang2023self}$^{\ast}$& \checkmark & 72.1	& 85.9 &	65.0	& 65.8 &	85.8 &	61.1 &	63.8 &	73.2	& 56.1 &	71.6 &	75.3 &	85.7 &	71.8\\
      G-SFDA (ViT-B)$^{\ast}$~\cite{yang2021generalized}& \checkmark & 71.6 &	85.4 &	64.9 &	67.1 &	83.9 &	62.1 &	64.2	& 73.8	& 57.0 &	72.3 &	75.8	& 86.0 &	72.0 \\
      Mixup~(ViT-B)$^{\ast}$~\cite{kundu2022balancing}  & \checkmark & \textbf{74.3} &	86.1 &	65.2 &	66.3 &	86.3 &	61.8 &	64.0	& \textbf{74.1}	& 56.8	& 72.1	& 74.9	& 86.1 &	72.3\\
      SHOT++(ViT-B)$^{\ast}$~\cite{liang2021source} & \checkmark  & 73.3 &	85.2	& 67.8 &	64.6 &	\textbf{87.4}	& 61.9	& 64.7 &	73.3 &	57.6	& 72.2	& 73.4	& \textbf{88.3}	& 72.5 \\
      Improved SFDA~(ViT-B)$^{\ast}$~\cite{mitsuzumi2024understanding}  & $\checkmark$ & 71.1 &	\textbf{86.8} &	66.1	& 67.8	& 86.3 &	63.1	& 64.2 &	72.8 &	56.3 &	72.4 &	\textbf{76.1} &	88.2 &	72.6\\
      \textbf{CADTrans~(ViT-B)} & \checkmark &70.7 &	82.3	& \textbf{68.0} &	\textbf{71.7} &	85.1 &	\textbf{67.8} &	68.0 &	72.7 &	\textbf{61.4}	& \textbf{74.0} &	73.6 &	82.7	& \textbf{73.2}\\
      \midrule
      DieT-B~(SO)~\cite{touvron2021training} & -- & 61.3	& 71.4	& 60.4 &	65.2	& 81.3	& 58.6 &	62.8 &	69.7	& 54.1 &	67.4	& 65.5	& 73.1 &	65.9 \\
      DIPE (DeiT-B)$^{\ast}$~\cite{wang2022exploring} & \checkmark & 70.2 &	81.3 &	66.5 &	70.7	& 83.7	& 65.6	& 67.5 &	71.4	 & 60.8	& 71.6 &	71.6	&	 79.9 &	71.7 \\
      G-SFDA (Deit-B)$^{\ast}$~\cite{yang2021generalized} & \checkmark & 71.9	& 79.8	& 64.5	&  71.0 & 	83.2 & 	65.6	&  64.3	&  74.2	&  57.6	&  72.6	&  76.3 & 	87.6	& 72.4\\
      Mixup~(DeiT)$^{\ast}$~\cite{kundu2022balancing} & \checkmark & 71.6 &	80.3 &	65.4 &	\textbf{71.6} &	81.0 &	66.4 &	\textbf{69.3} &	67.9	& 63.4 &	72.3	& \textbf{76.4}	& 84.5	& 72.5\\
      SHOT++(DeiT-B)$^{\ast}$~\cite{liang2021source} & \checkmark &74.6&	88.2	& 66.8 &	66.6 &	88.4 &	62.8 &	63.7 &	74.3 &	57.8 &	71.5 &	75.4 &	87.3 &	73.1 \\
      Improved SFDA~(ViT-B)$^{\ast}$~\cite{mitsuzumi2024understanding}  & $\checkmark$ & 74.9 &	89.0 &	\textbf{66.9}	& 68.1	& 87.1 &	62.6 &	63.9 &	73.1 &	57.5 &	73.0	& 75.8	& 88.1	& 73.3 \\
      \textbf{CADTrans~(DeiT-B)} & \checkmark  & 
      \textbf{75.4} &	\textbf{89.2} &	66.5 &	68.0 &	\textbf{89.3} &	63.1 &	64.1	& \textbf{74.5}	& 58.4 &	\textbf{73.5}	& 76.3 &	\textbf{88.9}	& \textbf{73.9} \\
      \Xhline{1.1pt}
  \end{tabular}
  }
  \vspace{-4mm}
\end{table*}

\textbf{Results of Office-Home.} 
According to the results on the Office-Home dataset shown in Table~\ref{tab:office-home}, 
{ the traditional ResNet model only achieved a score of 46.1\% on the Office-Home dataset.
SHOT, G-SFDA, and SHOT++ are all utilizing pseudo-label evaluation for self-supervised learning, with SHOT++ obtaining the best score of 73.0\%.
Dac and SFDA-DE exploit MMD to align the proxy source domain, they achieve the similar score of pseudo-label evaluation. 
CRMA exploit consistency regularization strategies of augmented sample to achieve the alignment of target domain, which obtains the best performance of 73.9\%. 
In comparison, our multi-consistency strategies with assistant domain alignment can overcome other approaches and obtain the SOTA performance of 74.1\%.
}
For ViT-Based approaches, directly exploiting the ViT-B to evaluate the target domain benchmark before adaptation~(i.e., Source-only) can obtain a score of 73.6\%. In which TransDA first exploit the ViT-based backbone to solve the SFDA task, which can achieve  an intermediate score of 79.3\%.
In comparison, the performance of our CADTrans~(ViT-B) is 81.7\% in the source-free setting, surpassing the TransDA and SHOT++~(ViT-B) by 2.4\% and 1.5\% respectively, which proves the necessity of exploiting the consistent strategy of global attention in domain adaptation.
For DeiT-based approaches, SHOT++~(DeiT) and G-SFDA~(DeiT), both utilizing the DeiT-B backbone, achieve average scores of 77.0\% and 80.5\%, respectively. Currently, DSiT-B addresses the SFDA challenge from both domain-specific and task-specific perspectives, achieving an average score of 80.5\%. In contrast, our CADTrans~(DeiT) model, also implemented with the DeiT-B backbone, reaches a top score of 82.9\%, surpassing the state-of-the-art performance by 1.4\%.
Additionally, both of them achieve the highest scores in multiple sub-domains such as Cl $\rightarrow$ Ar, Cl $\rightarrow$ Pr, Rw $\rightarrow$ Pr, etc.

\textbf{Results of VisDA-C.} 
As shown in Table~\ref{tab:office-31}, we mainly compare the results of SFDA on the ResNet-101 backbones due to its super performance.
Firstly, we observe that the `Source-only' approach implemented with DeiT-B/ViT-B outperforms the one using the ResNet-101 backbone by 25.1\% and 27.5\%, respectively.
This shows the superiority of transformers on DA.
Therefore, for CNN-based methods, a more rational design is needed. 
The best performance of source-free ResNet-101 approach CR-SFDA on the VisDA-C dataset achieves an average accuracy of 88.7\%, which outperforms the other results of NRC, SFDA-DE, SHOT++, and Improved SFDA. Our approach is slightly lower than CR-SFDA and Improved SFDA methods. 
For ViT-base approaches, the baseline of Source-only is 66.1\%. Most of the methods we implemented scored around 80\%, with SHOT++ achieving a best score of 87.6\%.
Our CADTrans (ViT-B) can outperform the others to achieve the SOTA performance of 87.8\%.
Likewise, CADTrans (DeiT-B) can also outperforms the other benchmarks by 0.6\%.
It is worth noting that this increase appears relatively minor, mainly because averaging dampens significant performance in certain subdomains.
Finally, we argue that our approach can extract additional information from the assistant domain to compensate for the invariable feature representation while our consistency strategy based on the assistant domain can provide some improvement for the progress of domain adaptation. 
\begin{table}[!htbp]
  \centering
  \caption{Effect of ADM block for Supervised training and `Source-only' adaptation in the source domain.}
  \resizebox{0.8\linewidth}{!}{
    \begin{tabular}{ccccc}
    \Xhline{1.1pt}
    \multirow{2}[4]{*}{DataSets} & \multicolumn{2}{c}{Supervised} & \multicolumn{2}{c}{Source-only} \\
    \cmidrule{2-5}          & ViT-B & CADTrans & ViT-B & CADTrans \\
    \midrule
    Office-31 & 95.9  & 96.5  & 84.6  & 86.2 \\
    Office-Home & 86.5  & 89.3  & 72.7  & 73.6 \\
    VISDA-C & 98.5  & 99.5  & 64.9  & 66.1 \\
    \midrule
    Avg.  & 93.6  & 95.1  & 74.1  & 75.3 \\
    \Xhline{1.1pt}
    \end{tabular}%
   }
   \vspace{-2mm}
   \label{tab:SD_ADM}%
\end{table}%

\textbf{Results of DomainNet-126.} 
  To further substantiate the efficacy of our approach in large-scale domain adaptation, we conduct additional experiments on the DomainNet~\cite{peng2019moment} benchmark and compare the results with various baselines.
  As demonstrated in Table~\ref{tab:fourdn}, utilizing ResNet-50 as the backbone network yields an average accuracy score of only 58.1\% in the 'Source-only' setting.
  Approaches built upon ResNet-50, including  AdaCon, PLUE, TPDS, and NRC, can enhance the top performance to 67.5\%. While SHOT, CoWA, GKD, and Improved SFDA are both achieve the performance around 68\%. Thanks to auxiliary domains and consistency strategies, our CADRN method outperforms the GKD and Improved SFDA by 1.2\%.
  In the 'Source-only' setting, ViT-B and DeiT-B achieve baseline performances of 65.2\% and 65.9\%, respectively. Most ViT-based approaches can surpass 70\% performance.
  This proves that the models of the transformer-based architecture with self-attention have significant advantages on large-scale DA tasks.
  Leveraging an enhanced architecture that incorporates domain consistency and CMK-MMD, our approaches built upon variants of CADTrans (ViT-B) and CADTrans (DeiT-B) outperform the SOTA benchmarks by {0.6} points, respectively, thereby attaining the highest performance.

\begin{figure*}[!ht]
  \centering
  \includegraphics[width=0.85\linewidth]{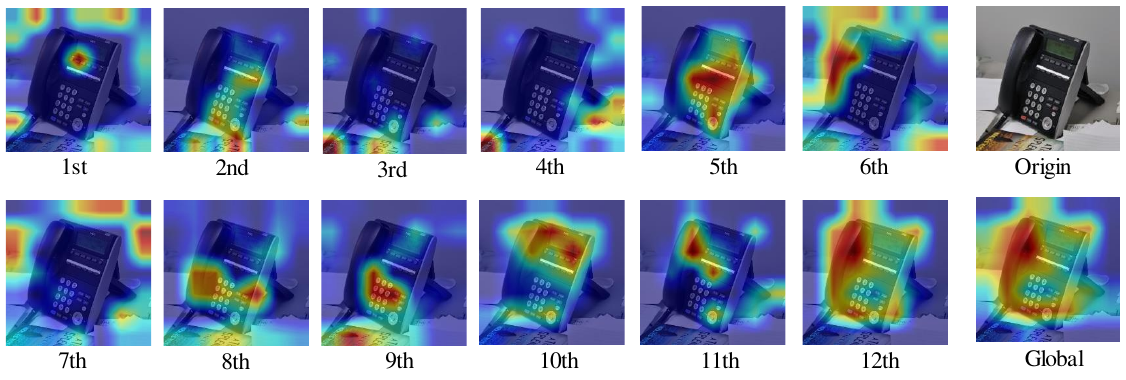}
  \caption{Attention maps of the intermediate layers in CADTrans~(ViT-B) model. The \textbf{Right:} the right of picture is attention map of each layer. The \textbf{Left:} the left of picture are original image of the sample and the final global attention map aggregated by our approaches.}
  \label{fig:multilevel_attention}
  \vspace{-2mm}
\end{figure*}

\subsection{Ablation study}
\textbf{Effect of ADM on source domain.}
Since the ADM block is introduced for our CADTrans. Thus, we first validate the effect of ADM block for supervised training and `Source-only' adaptation in the source domain, as shown in Tab.~\ref{tab:SD_ADM}. 
We verify the effectiveness of the ADM block from two aspects. First, we verify the model by pure ViT-B~(i.e., w/o ADM ) and our CADTrans with the distilled training~(i.e., w ADM). 
When exploiting the pure ViT-B model to train both three datasets by supervised learning. In the Office-31 dataset, we can obtain an average score of 95.9\%, while exploiting CADTrans i.e., adding the ADM block to the ViT-B model, we can achieve 96.5\% on average, roughly a 0.6\% improvement. On both three datasets, CADTrans can improve the average result of the model by 1.5\%.
Based on the above two forms of architecture, we apply them to the target domain without exploiting any adaptation strategies, i.e., `Source-only' adaptation.  
The score of `Source-only' by CADTrans on three datasets can significantly be improved by 1.2\% on average.
We believe CADTrans enhances the ability of ViT to learn the spatial context from the global features, thus enhancing the inductive bias to some extent.
\begin{table}[!htbp]
  \centering
  \caption{The ablation study of our approaches exploited by ViT-backbone with different components. `+' denotes the add operation.}
  \resizebox{0.95\linewidth}{!}{
    \begin{tabular}{lcccc}
    \Xhline{1.1pt}
    Method &  Office-31 & Office-Home & VISDA-C & Avg. \\
    \midrule
    Source-only & 86.2 & 73.6 &  66.1 & 75.3 \\
    Baseline & 91.3 &  79.4  & 85.6  & 85.4 \\
    + $\mathcal{L}_{cst}$   & 92.5 & 81.1  & 86.7 & 86.8 \\
    + $\mathcal{L}_{cst}$ + $\mathcal{L}_{cmk}$  & 93.0 &  81.7  &  87.8 & 87.5 \\
    \Xhline{1.1pt}
    \end{tabular}%
  \label{tab:ablation_study}%
  \vspace{-4mm}
  }
\end{table}%

\textbf{Effect of strategies.} 
To validate the effectiveness of the different components of our method, we exploit the ViT-B backbone to perform ablation experiments on the three datasets for each component.
The corresponding results are reported in Tab.~\ref{tab:ablation_study}.
In the first row, training the original model on the source domain without any adaptation, we obtain 86.2\%, 73.6\%, and 66.1\%.
If we directly utilize the information maximization loss $\mathcal{L}_{im}$ to optimize the model~(i.e., Baseline), we can achieve 91.3\%, 79.4\%, and 85.6\%, respectively.
Moreover, by utilizing the $\mathcal{L}_{cst}$ to supervise the alignment of the source and target domain, the average result can be improved by 1.4\%, which proves that our dynamic consistency strategy can effectively distinguish inconvenient features, so as to distinguish between easy samples and hard samples and improve the effect of the model through multiple consistency.
Finally, both $\mathcal{L}_{cst}$ and $\mathcal{L}_{cmk}$ are employed to verify our approach, which improves the baseline by about 2.1\%. This demonstrates that both the proposed three components are critical for CADTrans to perform well on SFDA.

\subsection{Visualization and analysis}
To explore the alignment effect of the final source domain and target domain, we conduct visualization experiments from two aspects: the attention maps by grad-cam~\cite{jacobgilpytorchcam} and feature alignment by t-SNE.
\begin{figure}[!htbp]
  \centering
  \includegraphics[width=\linewidth]{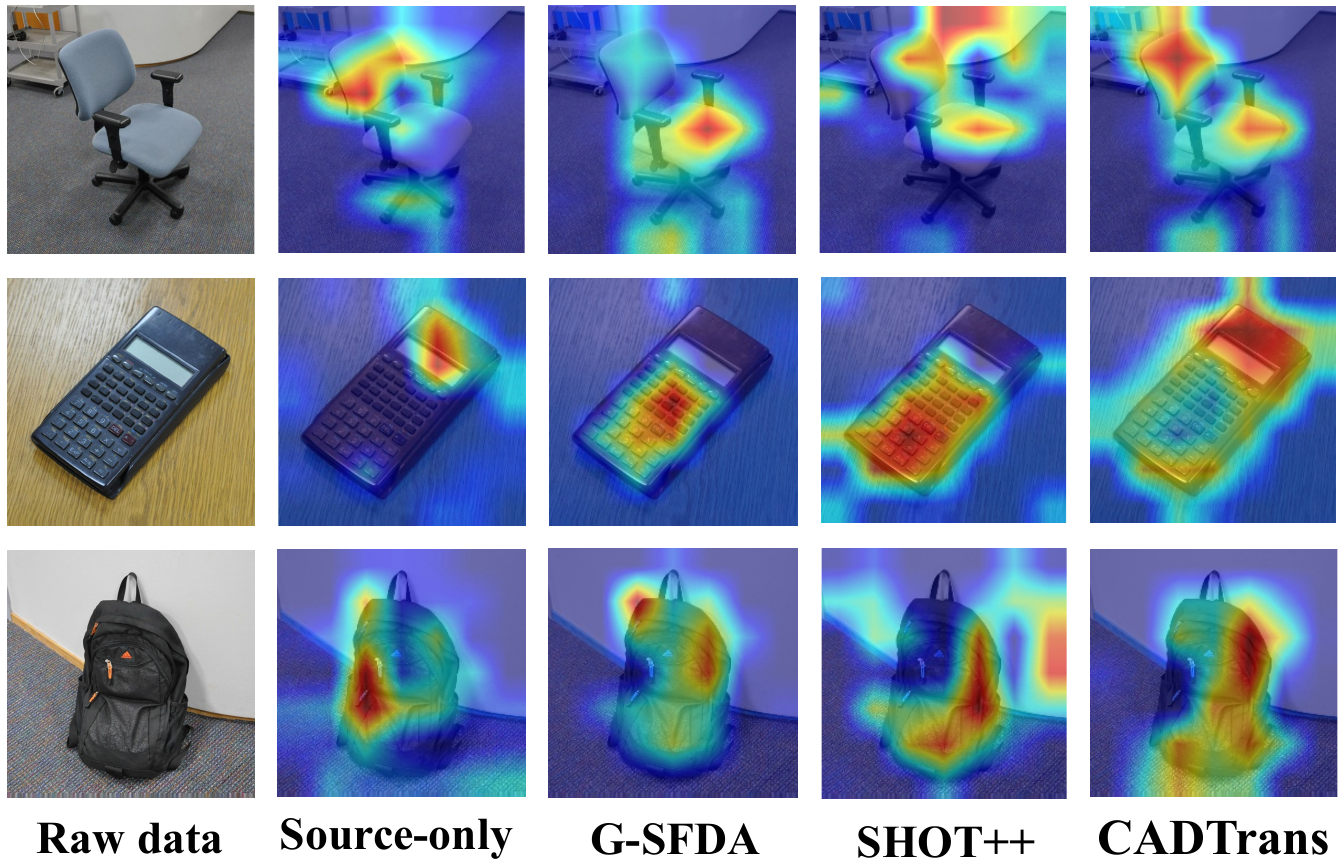}
  \vspace{-4mm}
  \caption{Attention maps of images about desk chair, calculator, and black package in the Office-31 dataset. }
  \label{fig:attention}
\end{figure}

\textbf{Attention Map}. In the ViT model, the attention maps represent the degree of attention to the target region. Normally, the hotter the color of the area, the higher the attention to the area.
According to attention maps shown in Figure~\ref{fig:attention}, the original `Source-only' can hardly pay attention to the main regions. 
Although G-SFDA~(ViT-B) and SHOT++~(ViT-B) can focus on the main objects, it is not comprehensive enough.
Compared with the former methods, our proposed method CADTrans can accurately capture discriminative region features.
For example, the `Source only' method only pays attention to the background of the images w.r.t. calculator, while our proposed method focuses on most areas of the target, which proves that the attention effect was effectively improved after alignment.

\begin{figure}
  \centering
  \vspace{-2mm}
   \subfloat[Before Adaptation]{
    \includegraphics[width=0.45\linewidth]{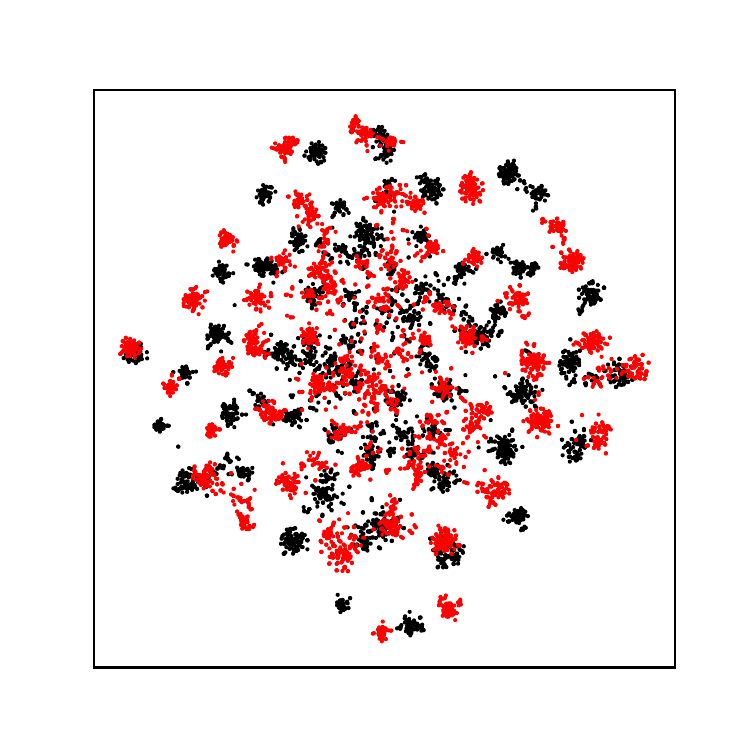}
    \label{fig:tsne_a}
   }
   \subfloat[After Adaptation]{
    \includegraphics[width=0.45\linewidth]{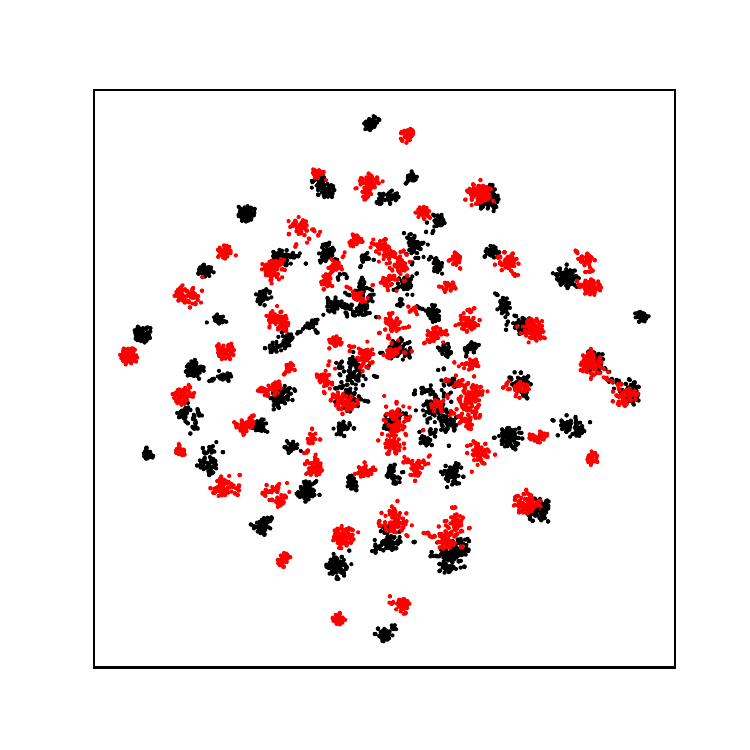}
    \label{fig:tsne_b}
   }
   \caption{t-SNE visualization for domain adaptation on Office-Home (A $\rightarrow$ P), Red denotes the source domains, while black denotes target domains.}
  \label{fig:tsne_ab}
\end{figure}

\textbf{t-SNE}. To demonstrate the effect of different methods on domain alignment, we utilize t-SNE to visualize the distributions of feature representations, which are obtained from the penultimate layer in both source and target domains.
As can be seen from Figure~\ref{fig:tsne_a}, the sample features of the same category are more dispersed before adaptation.
This might be due to the severe domain shift problem with source data.
Benefiting from $\mathcal{L}_{cst}$ and $\mathcal{L}_{cmk}$ loss, after adaptation, we can observe that the category distance is significantly reduced and the sample categories are distributed more clearly.

\subsection{Effect of hyper-parameters}
\begin{figure}[!htbp]
  \centering
  \vspace{-2mm}

   \subfloat[$\alpha$]{
    \includegraphics[width=0.46\linewidth]{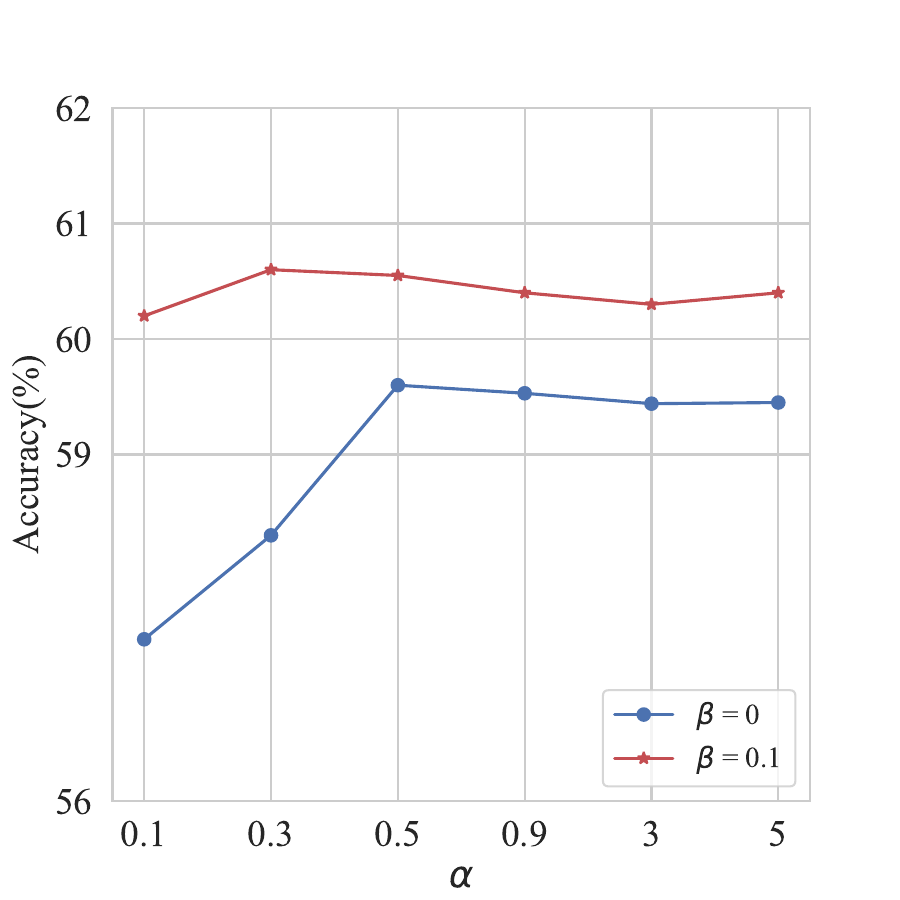}
      \label{fig:param_a}
   }
   \hfill
   \subfloat[$\beta$]{
    \includegraphics[width=0.46\linewidth]{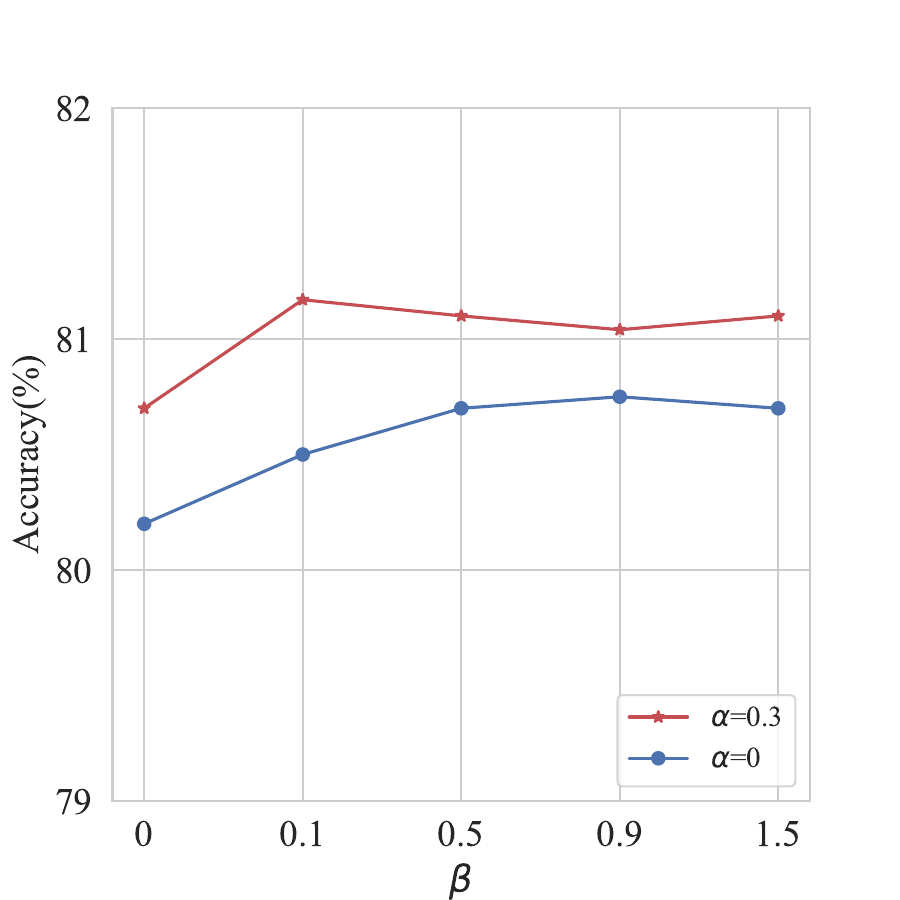}
      \label{fig:param_b}
   }
  \vspace{-2mm}
  \caption{{Sensitivity of hyper-parameter  $\alpha$ and $\beta$ in specific tasks~(Ar$\rightarrow$Cl and Pr$\rightarrow$Ar) on Office-Home Dataset.} }
  \label{fig:param_ab}
\end{figure}

In our total loss function (Eq.~\ref{eq:total_loss}), $\alpha$ and $\beta$ are the major hyper-parameters for balancing the loss terms in our framework.
To test their effect on the final performance, we conduct an experiment on the following two tasks, Ar$\rightarrow$Cl and Pr$\rightarrow$Ar.
As depicted in Figure~\ref{fig:param_a}, our model is less sensitive to the change of $\alpha$ and  $\beta$, and the results are significantly improved when $\alpha$ and $\beta$ are larger than 0.
For the Ar$\rightarrow$Cl task, when $\beta$ is set to 0.1, the discrepancy of $\alpha$ over the interval is only about 0.5\%.
While for the task Pr$\rightarrow$Ar, the best performance is achieved when $\beta$ is set to 0.1 and $\alpha$ is set to 0.3, as shown in Figure~\ref{fig:param_b}.

\subsection{Mechanisms and architectural details of ADM}
To mitigate the effect of lacking inductive bias and obtain invariant feature representations, we make full use of the advantages of CNN to compensate for this deficiency. Therefore, we construct an ADM black to extract the feature representations $\hat{f}_{a}$ and logits $\hat{z}_{a}$ from the aggregated multilevel global features {$\hat{x}_{a} \in \mathcal{R}^{ \mathrm{(N+1)}\times \mathrm{D}} $} of intermediate layers. We first reshape the self-attention to feature map $\hat{x}_{a} \in \mathcal{R}^{ \mathrm{D} \times \mathrm{H}\times \mathrm{W}}$, {where D is dimension of latent vector as 768 in ViT-B/16, (N+1) donates the sequence length of positional tokens and CLS token, which is obtained by H$\times$W+1}.  In our reshaped map, the latent vector can be seen as a channel-wise feature, therefore,  we exploit the convolution to downsample the feature map.  As can be seen from Tab.~\ref{tab:adm_arch}, {our ADM of CADTrans framework consists of three 3$\times$3 convolution layers.}  By downsampling operation, we can obtain the related context of attention features, which covers the major object. For example, the ViT-B model takes the $768$ dimensional features as a hidden vector. To make the dimension consistent with the classifier, we extract $256$ dimensional features as $\hat{f}_{a} \in \mathcal{R}^{\mathrm{B} \times 256} $  and map the features to logits as $\hat{z}_{a}\in \mathcal{R}^{ \mathrm{B} \times \mathrm{C}} $ by the FC layer which consists of two linear layers. This module can be applied dynamically in CADTrans~(ViT-B) and CADTrans~(DeiT-B) models to assist the model generate the assistant domain. 
To further adapt our method to traditional CNN models, we integrated the ADM into the ResNet architecture. First, we extracted features from each layer of ResNet separately, resulting in four different dimensional feature representations. For instance, the dimensions of the four feature representations in the middle layer of ResNet50 are $\hat{f}_{1}^{a}\in \mathcal{R}^{B \times 256 \times 56 \times 56}$ ,  $\hat{f}_{2}^{a}\in \mathcal{R}^{B \times 512 \times 28 \times 28}$,  $\hat{f}_{3}^{a}\in \mathcal{R}^{B \times 1024 \times 14 \times 14}$, and $\hat{f}_{4}^{a}\in \mathcal{R}^{B \times 2048 \times 7 \times 7}$. We then applied a 1$\times$1 convolution to reduce and align the features in the channel dimensions as 256, followed by interpolation to ensure consistency across dimensions as $\hat{f}^{a} \in \mathcal{R}^{ B \times 256 \times 56 \times 56} $. This fusion is similar to spatial pyramid fusion~\cite{dong2023feature}. Finally, we obtained global features $\hat{x}_{a}$ by performing an aggregation using an exponential moving average, which is subsequently used as inputs to the ADM model as depicted in following CADRN of the Tab~\ref{tab:adm_arch}.

\begin{table}[!htbp]  
  \centering
  \caption{The detailed architecture of ADM block. Where Fea. Dim. stands for Feature Dimensions.}
  \setlength\extrarowheight{1pt}
  \resizebox{\linewidth}{!}{
    \begin{tabular}{c|cc}
    \Xhline{1.1pt}
    Arch. & Input and Output of Fea. Dim. &  Components \\
    \hline
    \multirow{7}{*}{CADTrans} &[B, 768, 14, 14]$\rightarrow$[B, 512, 6, 6]  &  Conv(3$\times$3, $\mathrm{C}_{o}$=512, S=2),   \\ 
    \multirow{7}{*}{}  &  &  BN, ReLU    \\ 
    
     \multirow{7}{*}{} &[B, 512, 6, 6] $\rightarrow$ [B, 256, 4, 4] & Conv(3$\times$3, $\mathrm{C}_{o}$=256, S=1),  \\
     \multirow{7}{*}{} &  & BN, ReLU   \\

     \multirow{7}{*}{} &[B, 256, 4, 4] $\rightarrow$ [B, 256, 2, 2] &  Conv(3$\times $3,  $\mathrm{C}_{o}$=256, S=1),   \\
    \cline{2-3}
    \multirow{7}{*}{} &[B, 256, 2, 2] $\rightarrow$ [B, 256]  & \quad AveragePool, Faltten   \\
    
    \multirow{7}{*}{} &[B, 256 ] $\rightarrow$ [B, C] &  \quad  Linear, ReLU, Linear  \\
    
    \Xhline{1.1pt}

    \multirow{7}{*}{CADRN} & [B, 256, 56, 56]  $\rightarrow$ [B, 256, 28, 28]  &  Conv(1$\times $1, $\mathrm{C}_{o}$=256, S=2) \\ 
     &  & BN, ReLU  \\ 
    
    \multirow{7}{*}{} & [B, 256, 28, 28] $\rightarrow$ [B, 256, 14, 14] &  Conv(1$\times $1, $\mathrm{C}_{o}$=256, S=2)  \\
    &  & BN, ReLU \\

    \multirow{7}{*}{} &[B, 256, 14, 14] $\rightarrow$ [B, 256, 7, 7]  & Conv(1$\times $1,  $\mathrm{C}_{o}$=256, S=2),  \\
    \cline{2-3}
    \multirow{7}{*}{} & [B, 256, 7, 7] $\rightarrow$ [B, 256]  & \quad AveragePool, Faltten  \\
    
    \multirow{7}{*}{} & [B, 256 ] $\rightarrow$ [B, C] &  \quad  Linear, ReLU, Linear  \\
    \Xhline{1.1pt}
    \multicolumn{3}{l}{\small * S is the stride of the Conv2D layer. } \\
    \multicolumn{3}{l}{\small * $\mathrm{C}_{o}$ is the output dimension of the linear layer.}
    \label{tab:adm_arch}  
    \end{tabular}
    }
  \vspace{-4mm}
  \end{table}

\subsection{Computational overhead evaluation}
Our proposed method introduces ADM and develops a new backbone model. To thoroughly assess the model's computational overhead and ensure it meets the requirements for real-world applications, we evaluate it from two perspectives: the amount of computations, measured in floating point operations per second~(FLOPs), and the number of parameters~(Params). The results of this evaluation are presented in the Tab.~\ref{tab:overhead} below. As can be seen from the table, the overhead of our approach is not very high, and the introduced modules only add a maximum of 0.1$\times$ overhead in the FLOPs evaluation by CADRN-50. Furthermore, all of the improved architectures result in an increase of fewer than 0.1$\times$ in the number of parameters. 
This suggests our method can be applied effectively in practice, equating it to the original model.
\begin{table}[htbp] 
  \centering
  \caption{A comparison of the computational overhead between our reconstructed models and the original models.}
  \resizebox{0.95\linewidth}{!}{
    \begin{tabular}{ccccc}
    \toprule
    Ours / Orignal & FLOPs (G) & Rate & Params (M) & Rate \\
    \midrule
    CADRN-50/RN-50 & 4.59 / 4.13 & $\sim$1.11$\times$  & 24.76 / 23.51 & $\sim$1.05$\times$ \\
    CADRN-101/RN-101 &  8.32 / 7.86 & $\sim$1.06$\times$ & 43.76 / 42.50 & $\sim$1.03$\times$ \\
    CADTrans(ViT-B)/ViT-B &  16.94 / 16.86 & $\sim$1.00$\times$ &  88.70 / 86.42 & $\sim$1.03$\times$ \\
    CADTrans(DeiT-B)/DeiT-B & 16.94 / 16.86 & $\sim$1.00$\times$ &  88.70 / 86.42 & $\sim$1.03$\times$ \\
    \bottomrule
    \end{tabular}%
  }
  \label{tab:overhead}%
  \vspace{-8mm}
\end{table}%

\subsection{Visualization of multilevel attention}
To study whether the target is noticed or not by attention, we visualize heatmap attention feature maps of each layer by the grad-cam and calculate the final global attention visualized features by EMA, as depicted in Fig.~\ref{fig:multilevel_attention}. 
From the right of Fig.~\ref{fig:multilevel_attention}, displaying the attention map for each layer, we can observe that the target objects are not completely covered by the attention maps in each layer. For example, in layers of 1st-4th, the target samples are barely focused on. And the others such as 5th-10th, etc., these layers can focus on different local features of targets. In the last 12th layer, the target object can be noticed to a large extent. 
Besides, we exploit the EMA to calculate the global attention maps, which can focus on the vast majority of the target regions and fully cover the front attention regions. 
This powerful visualization illustrates that the EMA method can effectively capture the global features of the object. 
Based on such feature maps, the ADM block can obtain more accurate feature representations.

\subsection{Limitations and future work}
 In this paper, we propose a consistent assistant domains transformer to solve the domain shift of SFDA. Extensive experiments demonstrate that our method performs well on most benchmark datasets. However, our method also has certain limitations as follows:
\begin{itemize}
  \item \textbf{Limitation of model pre-training}. In this paper, we propose to improve the backbone with an ADM block, which requires plugging the module into the original backbone before release, which may increase the complexity of using the model.
  \item \textbf{Fairness of assessment}. With the introduction of ADM, we can improve the model's training effectiveness from a holistic perspective. This may result in unfair evaluations of other baselines.
\end{itemize}

 Besides, in the task of SFDA, we notice there are still some difficulties that are urgent to be well addressed.
We leave these meaningful and thought-provoking challenges for future work and hope to further contribute to SFDA community by addressing these issues.
\begin{itemize}
  \item \textbf{Negative transfer}. Negative transfer is a significant challenge in DA, primarily caused by data-related issues. For instance, scene noise, such as variations in background and lighting within the target domain, can create inconsistencies between the source and target domains. This inconsistency leads to bias during the adaptation process. Therefore, in our future work, we will focus on enhancing data quality and finding effective ways to eliminate interference.
  \item \textbf{Model lightweighting}. Currently, the self-attention mechanism of the transformer model generates relatively high computational overhead. To address this, we plan to employ new model architectures (e.g., Mamba, Jamba, etc.) or compression techniques, such as distillation, quantization, and pruning, to reduce the model's computational complexity. Exploring additional ways to enhance the model's efficiency is an interesting avenue worth further investigation.
  \item \textbf{Extended to the new fields}. Besides, we are poised to broaden our research to encompass the advanced SFDA fields of large language models~(LLMs) and multimodal applications, aiming not only to reduce model weights but also to tackle the challenges of domain shifts effectively. By leveraging more advancements, we hope to significantly enhance the efficiency and performance of LLMs.
\end{itemize}

\section{Conclusion}
In this paper, we propose a novel approach CADTrans with a coincident assistant domain. The main innovations of CADTrans are attributed to three technical components: assistant domain, consistency strategies, and CMK-MMD. CADTrans with an assistant domain obtains discriminative representations from aggregated global attention, while multiple consistent strategies can dynamically evaluate the pseudo labels for self-supervised learning and distinguish source-like easy samples and target-specific hard samples for SFDA. Besides, the CMK-MMD is exploited to align the hard samples and easy samples. The experimental results demonstrate the effectiveness of our proposed approaches.

\newpage
\bibliographystyle{IEEEtran}
\bibliography{IEEEabrv,ref}

\begin{thebibliography}{10}
\providecommand{\url}[1]{#1}
\csname url@samestyle\endcsname
\providecommand{\newblock}{\relax}
\providecommand{\bibinfo}[2]{#2}
\providecommand{\BIBentrySTDinterwordspacing}{\spaceskip=0pt\relax}
\providecommand{\BIBentryALTinterwordstretchfactor}{4}
\providecommand{\BIBentryALTinterwordspacing}{\spaceskip=\fontdimen2\font plus
\BIBentryALTinterwordstretchfactor\fontdimen3\font minus \fontdimen4\font\relax}
\providecommand{\BIBforeignlanguage}[2]{{%
\expandafter\ifx\csname l@#1\endcsname\relax
\typeout{** WARNING: IEEEtran.bst: No hyphenation pattern has been}%
\typeout{** loaded for the language `#1'. Using the pattern for}%
\typeout{** the default language instead.}%
\else
\language=\csname l@#1\endcsname
\fi
#2}}
\providecommand{\BIBdecl}{\relax}
\BIBdecl

\bibitem{russakovsky2015imagenet}
O.~Russakovsky, J.~Deng, H.~Su, J.~Krause, S.~Satheesh, S.~Ma, Z.~Huang, A.~Karpathy, A.~Khosla, M.~Bernstein \emph{et~al.}, ``Imagenet large scale visual recognition challenge,'' \emph{Int. J. Comput. Vision}, vol. 115, pp. 211--252, 2015.

\bibitem{yang2022image}
S.~Yang, W.~Xiao, M.~Zhang, S.~Guo, J.~Zhao, and F.~Shen, ``Image data augmentation for deep learning: A survey,'' \emph{arXiv preprint arXiv:2204.08610}, 2022.

\bibitem{pan2009survey}
S.~J. Pan and Q.~Yang, ``A survey on transfer learning,'' \emph{:IEEE T. Knowl. Data En.}, vol.~22, no.~10, pp. 1345--1359, 2009.

\bibitem{ganin2015unsupervised}
Y.~Ganin and V.~Lempitsky, ``Unsupervised domain adaptation by backpropagation,'' in \emph{Int. Conf. Mach. Learn.}\hskip 1em plus 0.5em minus 0.4em\relax PMLR, 2015, pp. 1180--1189.

\bibitem{long2015learning}
M.~Long, Y.~Cao, J.~Wang, and M.~Jordan, ``Learning transferable features with deep adaptation networks,'' in \emph{Int. Conf. Mach. Learn.}, 2015, pp. 97--105.

\bibitem{su2024m}
J.~Su, B.~Wang, Z.~Fan, Y.~Zhang, L.-L. Zeng, H.~Shen, and D.~Hu, ``M 2 dc: A meta-learning framework for generalizable diagnostic classification of major depressive disorder,'' \emph{IEEE Trans. on Med. Imaging}, 2025.

\bibitem{goodfellow2014generative}
I.~Goodfellow, J.~Pouget-Abadie, M.~Mirza, B.~Xu, D.~Warde-Farley, S.~Ozair, A.~Courville, and Y.~Bengio, ``Generative adversarial nets,'' \emph{Adv. in Neural Inf. Process. Syst.}, vol.~27, 2014.

\bibitem{long2018conditional}
M.~Long, Z.~Cao, J.~Wang, and M.~I. Jordan, ``Conditional adversarial domain adaptation,'' \emph{Adv. in Neural Inf. Process. Syst.}, vol.~31, 2018.

\bibitem{su2021few}
J.~Su, H.~Shen, L.~Peng, and D.~Hu, ``Few-shot domain-adaptive anomaly detection for cross-site brain images,'' \emph{IEEE Trans. on Pattern Anal. and Mach. Intell.}, vol.~46, no.~3, pp. 1819--1835, 2024.

\bibitem{saito2018maximum}
K.~Saito, K.~Watanabe, Y.~Ushiku, and T.~Harada, ``Maximum classifier discrepancy for unsupervised domain adaptation,'' in \emph{Proc. of the IEEE/CVF Conf. Comput. Vis. Pattern Recognit.}, 2018, pp. 3723--3732.

\bibitem{li2020model}
R.~Li, Q.~Jiao, W.~Cao, H.-S. Wong, and S.~Wu, ``Model adaptation: Unsupervised domain adaptation without source data,'' in \emph{Proc. of the IEEE/CVF Conf. Comput. Vis. Pattern Recognit.}, 2020, pp. 9641--9650.

\bibitem{xia2021adaptive}
H.~Xia, H.~Zhao, and Z.~Ding, ``Adaptive adversarial network for source-free domain adaptation,'' in \emph{Proc. of the IEEE/CVF Int. Conf. Comput. Vis.}, 2021, pp. 9010--9019.

\bibitem{liang2020we}
J.~Liang, D.~Hu, and J.~Feng, ``Do we really need to access the source data? source hypothesis transfer for unsupervised domain adaptation,'' in \emph{Int. Conf. Mach. Learn.}, 2020, pp. 6028--6039.

\bibitem{yang2021generalized}
S.~Yang, Y.~Wang, J.~van~de Weijer, L.~Herranz, and S.~Jui, ``Generalized source-free domain adaptation,'' in \emph{Proc. of the IEEE/CVF Int. Conf. Comput. Vis.}, 2021, pp. 8978--8987.

\bibitem{hou2016unsupervised}
C.-A. Hou, Y.-H.~H. Tsai, Y.-R. Yeh, and Y.-C.~F. Wang, ``Unsupervised domain adaptation with label and structural consistency,'' \emph{IEEE Trans. on Image Process.}, vol.~25, no.~12, pp. 5552--5562, 2016.

\bibitem{ren2022multi}
C.-X. Ren, Y.-H. Liu, X.-W. Zhang, and K.-K. Huang, ``Multi-source unsupervised domain adaptation via pseudo target domain,'' \emph{IEEE Trans. on Image Process.}, vol.~31, pp. 2122--2135, 2022.

\bibitem{zhang2022divide}
Z.~Zhang, W.~Chen, H.~Cheng, Z.~Li, S.~Li, L.~Lin, and G.~Li, ``Divide and contrast: Source-free domain adaptation via adaptive contrastive learning,'' \emph{Adv. in Neural Inf. Process. Syst.}, vol.~35, pp. 5137--5149, 2022.

\bibitem{zhao2019learning}
H.~Zhao, R.~T. Des~Combes, K.~Zhang, and G.~Gordon, ``On learning invariant representations for domain adaptation,'' in \emph{Int. Conf. on Mach. Learn.}\hskip 1em plus 0.5em minus 0.4em\relax PMLR, 2019, pp. 7523--7532.

\bibitem{dosovitskiy2020image}
A.~Dosovitskiy, L.~Beyer, A.~Kolesnikov, D.~Weissenborn, X.~Zhai, T.~Unterthiner, M.~Dehghani, M.~Minderer, G.~Heigold, S.~Gelly \emph{et~al.}, ``An image is worth 16x16 words: Transformers for image recognition at scale,'' \emph{Int. Conf. Learn. Represent.}, 2021.

\bibitem{carion2020end}
N.~Carion, F.~Massa, G.~Synnaeve, N.~Usunier, A.~Kirillov, and S.~Zagoruyko, ``End-to-end object detection with transformers,'' in \emph{Proc. of the Eur. Conf. Comput. Vis.}\hskip 1em plus 0.5em minus 0.4em\relax Springer, 2020, pp. 213--229.

\bibitem{zheng2021rethinking}
S.~Zheng, J.~Lu, H.~Zhao, X.~Zhu, Z.~Luo, Y.~Wang, Y.~Fu, J.~Feng, T.~Xiang, P.~H. Torr \emph{et~al.}, ``Rethinking semantic segmentation from a sequence-to-sequence perspective with transformers,'' in \emph{Proc. of the IEEE/CVF Conf. Comput. Vis. Pattern Recognit.}, 2021, pp. 6881--6890.

\bibitem{sun2016return}
B.~Sun, J.~Feng, and K.~Saenko, ``Return of frustratingly easy domain adaptation,'' in \emph{Proc. of the AAAI Conf. Artif. Intell.}, vol.~30, no.~1, 2016.

\bibitem{flamary2016optimal}
R.~Flamary, N.~Courty, D.~Tuia, and A.~Rakotomamonjy, ``Optimal transport for domain adaptation,'' \emph{IEEE Trans. Pattern Anal. Mach. Intell}, vol.~1, 2016.

\bibitem{kang2019contrastive}
G.~Kang, L.~Jiang, Y.~Yang, and A.~G. Hauptmann, ``Contrastive adaptation network for unsupervised domain adaptation,'' in \emph{Proc. of the IEEE/CVF Conf. Comput. Vis. Pattern Recognit.}, 2019, pp. 4893--4902.

\bibitem{ganin2016domain}
Y.~Ganin, E.~Ustinova, H.~Ajakan, P.~Germain, H.~Larochelle, F.~Laviolette, M.~Marchand, and V.~Lempitsky, ``Domain-adversarial training of neural networks,'' \emph{The J. of Mach. Learn. Res.}, vol.~17, no.~1, pp. 2096--2030, 2016.

\bibitem{tzeng2017adversarial}
E.~Tzeng, J.~Hoffman, K.~Saenko, and T.~Darrell, ``Adversarial discriminative domain adaptation,'' in \emph{Proc. of the IEEE/CVF Conf. Comput. Vis. Pattern Recognit.}, 2017, pp. 7167--7176.

\bibitem{sanyal2023domain}
S.~Sanyal, A.~R. Asokan, S.~Bhambri, A.~Kulkarni, J.~N. Kundu, and R.~V. Babu, ``Domain-specificity inducing transformers for source-free domain adaptation,'' in \emph{Proc. of the IEEE/CVF Int. Conf. Comput. Vis.}, 2023, pp. 18\,928--18\,937.

\bibitem{vaswani2017attention}
A.~Vaswani, N.~Shazeer, N.~Parmar, J.~Uszkoreit, L.~Jones, A.~N. Gomez, {\L}.~Kaiser, and I.~Polosukhin, ``Attention is all you need,'' \emph{Adv. in Neural Inf. Process. Syst.}, vol.~30, 2017.

\bibitem{yang2023tvt}
J.~Yang, J.~Liu, N.~Xu, and J.~Huang, ``Tvt: Transferable vision transformer for unsupervised domain adaptation,'' in \emph{Proc. of the IEEE/CVF Winter Conf. on Appl. of Comput. Vis.}, 2023, pp. 520--530.

\bibitem{xu2021cdtrans}
T.~Xu, W.~Chen, P.~Wang, F.~Wang, H.~Li, and R.~Jin, ``Cdtrans: Cross-domain transformer for unsupervised domain adaptation,'' \emph{Int. Conf. Learn. Represent.}, 2022.

\bibitem{sun2022safe}
T.~Sun, C.~Lu, T.~Zhang, and H.~Ling, ``Safe self-refinement for transformer-based domain adaptation,'' in \emph{Proc. of the IEEE/CVF Conf. Comput. Vis. Pattern Recognit.}, 2022, pp. 7191--7200.

\bibitem{yang2023self}
G.~Yang, Z.~Zhong, M.~Ding, N.~Sebe, and E.~Ricci, ``Self-training transformer for source-free domain adaptation,'' \emph{Appl. Intell.}, vol.~53, no.~13, pp. 16\,560--16\,574, 2023.

\bibitem{luo2022domain}
Z.~Luo, X.~Zhang, S.~Lu, and S.~Yi, ``Domain consistency regularization for unsupervised multi-source domain adaptive classification,'' \emph{Pattern Recognit.}, vol. 132, p. 108955, 2022.

\bibitem{tang2023consistency}
L.~Tang, K.~Li, C.~He, Y.~Zhang, and X.~Li, ``Consistency regularization for generalizable source-free domain adaptation,'' in \emph{Proc. of the IEEE/CVF Int. Conf. Comput. Vis.}, 2023, pp. 4323--4333.

\bibitem{liu2024domain}
S.~Liu, Q.~Wang, H.~Fan, W.~Ren, B.~Fan, and Y.~Tang, ``Domain consistency representation learning for lifelong person re-identification,'' \emph{arXiv preprint arXiv:2409.19954v2}, 2025.

\bibitem{lu2024consistency}
S.~L{\"u}, Z.~Li, X.~Zhang, and J.~Li, ``Consistency regularization-based mutual alignment for source-free domain adaptation,'' \emph{Expert. Syst. Appl.}, vol. 241, p. 122577, 2024.

\bibitem{liu2025consistency}
Z.~Liu, C.~Cui, C.~Zhang, F.~Meng, S.~Gong, M.~Xi, and L.~Li, ``Consistency-guided multi-source-free domain adaptation,'' \emph{Eng. Appl. Artif. Intel.}, vol. 139, p. 109497, 2025.

\bibitem{ding2022source}
N.~Ding, Y.~Xu, Y.~Tang, C.~Xu, Y.~Wang, and D.~Tao, ``Source-free domain adaptation via distribution estimation,'' in \emph{Proc. of the IEEE/CVF Conf. Comput. Vis. Pattern Recognit.}, 2022, pp. 7212--7222.

\bibitem{chen2023cf}
M.~Chen, M.~Lin, K.~Li, Y.~Shen, Y.~Wu, F.~Chao, and R.~Ji, ``Cf-vit: A general coarse-to-fine method for vision transformer,'' in \emph{Proc. of the AAAI Conf. Artif. Intell.}, vol.~37, 2023, pp. 7042--7052.

\bibitem{liang2022not}
Y.~Liang, C.~Ge, Z.~Tong, Y.~Song, J.~Wang, and P.~Xie, ``Not all patches are what you need: Expediting vision transformers via token reorganizations,'' \emph{Int. Conf. Learn. Represent.}, 2022.

\bibitem{deng2009imagenet}
J.~Deng, W.~Dong, R.~Socher, L.-J. Li, K.~Li, and L.~Fei-Fei, ``Imagenet: A large-scale hierarchical image database,'' in \emph{Proc. of the IEEE/CVF Conf. Comput. Vis. Pattern Recognit.}\hskip 1em plus 0.5em minus 0.4em\relax Ieee, 2009, pp. 248--255.

\bibitem{sun2017revisiting}
C.~Sun, A.~Shrivastava, S.~Singh, and A.~Gupta, ``Revisiting unreasonable effectiveness of data in deep learning era,'' in \emph{Proc. of the IEEE/CVF Int. Conf. Comput. Vis.}, 2017, pp. 843--852.

\bibitem{liang2021source}
J.~Liang, D.~Hu, Y.~Wang, R.~He, and J.~Feng, ``Source data-absent unsupervised domain adaptation through hypothesis transfer and labeling transfer,'' \emph{IEEE Trans. Pattern Anal. Mach. Intell.}, vol.~44, no.~11, pp. 8602--8617, 2021.

\bibitem{wang2022exploring}
F.~Wang, Z.~Han, Y.~Gong, and Y.~Yin, ``Exploring domain-invariant parameters for source free domain adaptation,'' in \emph{Proc. of the IEEE/CVF Conf. Comput. Vis. Pattern Recognit.}, 2022, pp. 7151--7160.

\bibitem{long2017deep}
M.~Long, H.~Zhu, J.~Wang, and M.~I. Jordan, ``Deep transfer learning with joint adaptation networks,'' in \emph{Int. Conf. on Mach. Learn.}\hskip 1em plus 0.5em minus 0.4em\relax PMLR, 2017, pp. 2208--2217.

\bibitem{zhu2020deep}
Y.~Zhu, F.~Zhuang, J.~Wang, G.~Ke, J.~Chen, J.~Bian, H.~Xiong, and Q.~He, ``Deep subdomain adaptation network for image classification,'' \emph{IEEE Trans. on Neural Netw. and Learn. Syst.}, vol.~32, no.~4, pp. 1713--1722, 2020.

\bibitem{he2016deep}
K.~He, X.~Zhang, S.~Ren, and J.~Sun, ``Deep residual learning for image recognition,'' in \emph{Proc. of the IEEE/CVF Conf. Comput. Vis. Pattern Recognit.}, 2016, pp. 770--778.

\bibitem{yang2021exploiting}
S.~Yang, J.~van~de Weijer, L.~Herranz, S.~Jui \emph{et~al.}, ``Exploiting the intrinsic neighborhood structure for source-free domain adaptation,'' \emph{Adv. in Neural Inf. Process. Syst.}, vol.~34, pp. 29\,393--29\,405, 2021.

\bibitem{mitsuzumi2024understanding}
Y.~Mitsuzumi, A.~Kimura, and H.~Kashima, ``Understanding and improving source-free domain adaptation from a theoretical perspective,'' in \emph{Proc. of the IEEE/CVF Conf. on Comput. Vis. Pattern Recognit.}, 2024, pp. 28\,515--28\,524.

\bibitem{kundu2022balancing}
J.~N. Kundu, A.~R. Kulkarni, S.~Bhambri, D.~Mehta, S.~A. Kulkarni, V.~Jampani, and V.~B. Radhakrishnan, ``Balancing discriminability and transferability for source-free domain adaptation,'' in \emph{Int. Conf. on Mach. Learn.}\hskip 1em plus 0.5em minus 0.4em\relax PMLR, 2022, pp. 11\,710--11\,728.

\bibitem{touvron2021training}
H.~Touvron, M.~Cord, M.~Douze, F.~Massa, A.~Sablayrolles, and H.~J{\'e}gou, ``Training data-efficient image transformers \& distillation through attention,'' in \emph{Int. Conf. Mach. Learn.}\hskip 1em plus 0.5em minus 0.4em\relax PMLR, 2021, pp. 10\,347--10\,357.

\bibitem{saenko2010adapting}
K.~Saenko, B.~Kulis, M.~Fritz, and T.~Darrell, ``Adapting visual category models to new domains,'' in \emph{Proc. of the Eur. Conf. Comput. Vis.}\hskip 1em plus 0.5em minus 0.4em\relax Springer, 2010, pp. 213--226.

\bibitem{venkateswara2017deep}
H.~Venkateswara, J.~Eusebio, S.~Chakraborty, and S.~Panchanathan, ``Deep hashing network for unsupervised domain adaptation,'' in \emph{Proc. of the IEEE/CVF Conf. Comput. Vis. Pattern Recognit.}, 2017, pp. 5018--5027.

\bibitem{peng2017visda}
X.~Peng, B.~Usman, N.~Kaushik, J.~Hoffman, D.~Wang, and K.~Saenko, ``Visda: The visual domain adaptation challenge,'' \emph{arXiv preprint arXiv:1710.06924}, 2017.

\bibitem{peng2019moment}
X.~Peng, Q.~Bai, X.~Xia, Z.~Huang, K.~Saenko, and B.~Wang, ``Moment matching for multi-source domain adaptation,'' in \emph{Proc. of the IEEE/CVF Int. Conf. Comput. Vis.}, 2019, pp. 1406--1415.

\bibitem{saito2019semi}
K.~Saito, D.~Kim, S.~Sclaroff, T.~Darrell, and K.~Saenko, ``Semi-supervised domain adaptation via minimax entropy,'' in \emph{Proc. of the IEEE/CVF Int. Conf. Comput. Vis.}, 2019, pp. 8050--8058.

\bibitem{chen2022contrastive}
D.~Chen, D.~Wang, T.~Darrell, and S.~Ebrahimi, ``Contrastive test-time adaptation,'' in \emph{Proc. IEEE/CVF Conf. Comput. Vis. Pattern Recognit. (CVPR)}, 2022.

\bibitem{Litrico_2023}
M.~Litrico, A.~Del~Bue, and P.~Morerio, ``Guiding pseudo-labels with uncertainty estimation for source-free unsupervised domain adaptation,'' in \emph{Proc. IEEE/CVF Conf. Comput. Vis. Pattern Recognit. (CVPR)}, 2023.

\bibitem{tang2024source}
S.~Tang, A.~Chang, F.~Zhang, X.~Zhu, M.~Ye, and C.~Zhang, ``Source-free domain adaptation via target prediction distribution searching,'' \emph{Int. J. Comput. Vision}, vol. 132, no.~3, pp. 654--672, 2024.

\bibitem{yang2021nrc}
S.~Yang, J.~van~de Weijer, L.~Herranz, S.~Jui \emph{et~al.}, ``Exploiting the intrinsic neighborhood structure for source-free domain adaptation,'' in \emph{Adv. in Neural Inf. Process. Syst. (NeurIPS)}, 2021.

\bibitem{lee2022confidence}
J.~Lee, D.~Jung, J.~Yim, and S.~Yoon, ``Confidence score for source-free unsupervised domain adaptation,'' in \emph{Int. Conf. Mach. Learn. (ICML)}, 2022.

\bibitem{tang2021model}
S.~{Tang}, Y.~Shi, Z.~Ma, J.~Li, J.~Lyu, Q.~Li, and J.~Zhang, ``Model adaptation through hypothesis transfer with gradual knowledge distillation,'' in \emph{IEEE/RSJ Int. Conf. on Intell. Robots and Syst. (IROS)}, 2021.

\bibitem{jacobgilpytorchcam}
J.~Gildenblat and contributors, ``Pytorch library for cam methods,'' \url{https://github.com/jacobgil/pytorch-grad-cam}, 2021.

\bibitem{dong2023feature}
W.~Dong, Y.~Yang, J.~Qu, Y.~Li, Y.~Yang, and X.~Jia, ``Feature pyramid fusion network for hyperspectral pansharpening,'' \emph{IEEE Trans. on Neural Netw. and Learn. Syst.}, 2023.

\end{thebibliography}


{
\vspace{-1cm}
\begin{IEEEbiography}[
   {\includegraphics[width=1in,height=1.25in,clip,keepaspectratio]{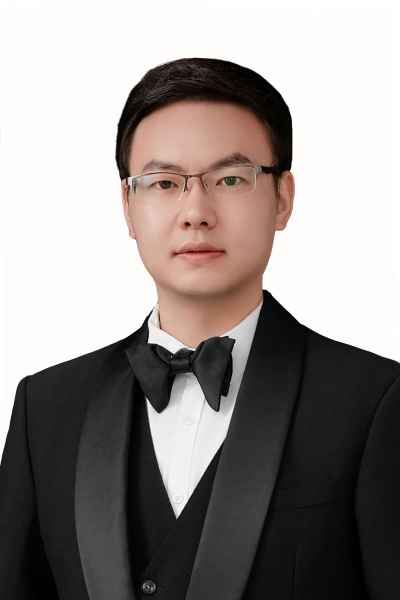}}]{Renrong Shao}
    received his Ph.D. degree in the School of Computer Science and Technology at East China Normal University in 2024. He is currently a lecturer and postdoctoral fellow in the Faculty of Military Health Services, Naval Medical University (Second Military Medical University), Shanghai, China. His research interests include computer vision, model compression, transfer learning, and intelligent healthcare.  
\end{IEEEbiography} 
}

{ 
\begin{IEEEbiography}[   
    {\includegraphics[width=1in,height=1.25in,clip,keepaspectratio]{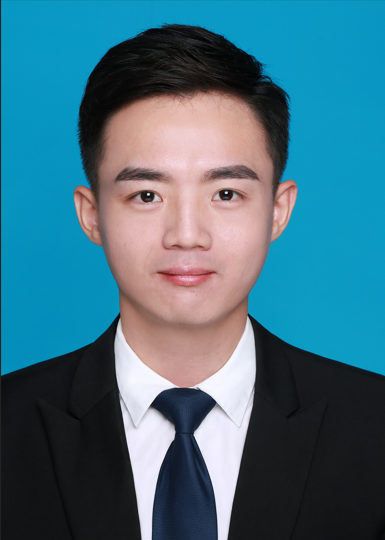}}]{Kangyang Luo} 
    obtained his PhD degree from the School of Data Science and Engineering at East China Normal University in 2024. He is currently a postdoctoral fellow in the Department of Computer Science and Technology at Tsinghua University, Beijing, China. His research interests include deep learning, federated learning, data-driven knowledge distillation, and large language models.
\end{IEEEbiography}
}

\newpage

{ 
\begin{IEEEbiography}[   
    {\includegraphics[width=1in,height=1.25in,clip,keepaspectratio]{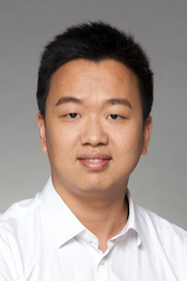}}]{Wei Zhang}(Member, IEEE) 
    received his Ph.D. degree in computer science and technology from Tsinghua University, Beijing, China, in 2016. He is currently a professor in the School of Computer Science and Technology, East China Normal University, Shanghai, China. His research interests mainly include user data mining and machine learning applications. He is a senior member of China Computer Federation.
\end{IEEEbiography}
}
{ 
\vspace{-11cm}
\begin{IEEEbiography}[   
    {\includegraphics[width=1in,height=1.25in,clip,keepaspectratio]{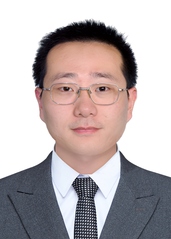}}]{Qin Li}(Member, IEEE) 
    received the Ph.D. degree from East China Normal University, in 2011. He is currently an Associate Professor with the
    Software Engineering Institute, East China Normal
    University, Shanghai, China. His research interests
    include trustworthy artificial intelligence and formal
    modeling and verification on self-adaptive intelligent systems.
\end{IEEEbiography}
}

{
\vspace{-11cm}
\begin{IEEEbiography}[
    {\includegraphics[width=1in,height=1.25in,clip,keepaspectratio]{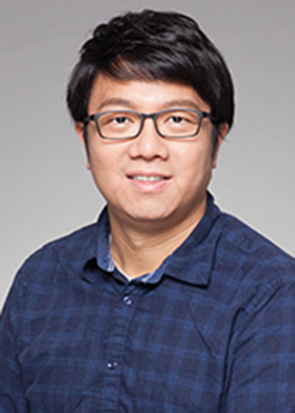}}]{Jun Wang} (Member, IEEE) received his Ph.D. degree in electrical engineering from Columbia University, New York, NY, USA, in 2011. Currently, he is a Professor at the School of Computer Science and Technology, East China Normal University, and an adjunct faculty member of Columbia University. From 2010 to 2014, he was a Research Staff Member at IBM T. J. Watson Research Center, Yorktown Heights, NY, USA. His research interests include machine learning, data mining, mobile intelligence, and computer vision. Dr. Wang has been the recipient of the Thousand Talents Plan in 2014.
\end{IEEEbiography}
}

\end{document}